\documentclass[preprint,1p,12pt]{elsarticle}
\usepackage{times}
\usepackage{nohyperref}
\usepackage{url}
\usepackage{graphicx,color}
\usepackage{epsfig}
\usepackage[algoruled,linesnumbered]{algorithm2e}
\usepackage{algorithmic}
\usepackage{amsmath}
\usepackage{amssymb}
\usepackage{array}
\usepackage{subfigure}
\usepackage{verbatim}
\usepackage{flushend}
\usepackage{booktabs}

\usepackage{amsmath,amsthm,amssymb}

\newtheorem{definition}{Definition}[section]

\newtheorem{remark}{Remark}[section]

\journal{Applied Energy}
\begin{document}
\begin{frontmatter}
\title{Energy Prediction using Spatiotemporal Pattern Networks}
%\tnotetext[mytitlenote]{The authors would like to thank Iowa Energy Center
%for supporting this work - opportunity grant no. OG-15-005.}
\author[]{Zhanhong Jiang$\dag$, Chao Liu$\dag$, Adedotun Akintayo$\dag$, Gregor P.\ Henze$\ddag$$\P$}
\author{Soumik Sarkar$\dag$\corref{corres}}
\ead{soumiks@iastate.edu}
\cortext[corres]{Corresponding author. Tel: +1 515 357 4328}
\address{$\dag$Department of Mechanical Engineering, 2025 Black Engineering\\
Iowa State University, Ames, IA 50011, USA\\
$\ddag$Department of Civil, Environmental and Architectural Engineering,\\
University of Colorado, Boulder, Colorado 80309, USA\\
$\P$National Renewable Energy Laboratory, Golden, CO 80401, USA}
%\ead{vchinde@iastate.edu}
%\author[address]{Venkatesh Chinde\corref{mycorrespondingauthor}}
%\ead{vchinde@iastate.edu, jheylmun@iastate.edu, adamkohl@iastate.edu,  zhjiang@iastate.edu, soumiks@iastate.edu,  akelkar@iastate.edu}
%\address[mymainaddress]{Department of Mechanical Engineering, 2038 Black Engineering\\
%Iowa State University, Ames, IA 50010, USA}
%\address[mysecondaryaddress]{Iowa State University, Ames, IA 50010, USA}

\begin{abstract}
This paper presents a novel data-driven technique based on the spatiotemporal pattern network (STPN) for energy/power prediction for complex dynamical systems. Built on symbolic dynamic filtering, the STPN framework is used to capture not only the individual system characteristics but also the pair-wise causal dependencies among different sub-systems. For quantifying the causal dependency, a mutual information based metric is presented. An energy prediction approach is subsequently proposed based on the STPN framework. For validating the proposed scheme, two case studies are presented, one involving wind turbine power prediction (supply side energy) using the Western Wind Integration data set generated by the National Renewable Energy Laboratory (NREL) for identifying the spatiotemporal characteristics, and the other, residential electric energy disaggregation (demand side energy) using the Building America 2010 data set from NREL for exploring the temporal features. In the energy disaggregation context, convex programming techniques beyond the STPN framework are developed and applied to achieve improved disaggregation performance.

%This paper presents a data-driven modeling framework to understand spatiotemporal interactions among wind turbines in a large scale wind energy farm as well as A recently developed probabilistic graphical modeling scheme, namely the spatiotemporal pattern network (STPN) is used to capture individual turbine characteristics as well as pair-wise causal dependencies. The causal dependency is quantified by a mutual information based metric and it has been shown that it efficiently and correctly captures both temporal and spatial characteristics of wind turbines. The causal interaction models are also used for predicting wind power production by one wind turbine using observations from another turbine. The proposed tools are validated using the Western Wind Integration data set from the National Renewable Energy Laboratory (NREL).
\end{abstract}
\end{frontmatter}

\section{Introduction}\label{sec:intro}
Energy prediction problems are essential for operating, monitoring, and optimizing (efficiency, cost) in diverse energy systems, from the supply side (e.g., wind energy, solar energy, power systems, battery) to the demand side (e.g., load monitoring, usage of electric vehicles, building energy management). Numerous studies are being carried out in terms of predicting the energy generation/consumption using time-series data \cite{ziel2016forecasting,liu2015highly,alessandrini2015analog,zuluaga2015short,wang2015study,garshasbi2016hybrid}. For instance, Kalman filtering, wavelet packet transforms, and least square support vector machines are used to predict wind power performance \cite{zuluaga2015short,wang2015study}, while an analog ensemble method is applied to forecast solar power \cite{alessandrini2015analog}. Liu et al. \cite{liu2015highly} predicts remaining state of charge of electric vehicle batteries based on predictive control theory. Hybrid genetic algorithms and Monte Carlo simulation approaches are applied to predict energy generation and consumption in net-zero energy buildings \cite{garshasbi2016hybrid}. For modern energy systems, a large number of subsystems is usually involved, for example, hundreds of wind turbines are closely collocated in a wind farm where the wind resource is similar and the conditions of them are analogous in terms of the power transmission to the power system. As a result, prediction of wind turbine output is related among each of them, and the characteristics of spatial interactions can be potentially applied for prediction \cite{jiang2015understanding} and design optimization. The prediction approaches discussed above can be viewed as methods of exploring temporal relationships. Spatial and temporal relationship widely exists in energy systems \cite{jain2014forecasting,liu2010prediction,jung2014current,kwon2010uncertainty}, yet spatiotemporal features are less commonly leveraged for energy prediction problems. The exploration of such features has been shown efficient in wind speed forecasting problems \cite{tascikaraoglu2016exploiting,jung2014current,tascikaraoglucompressive}.

To facilitate the energy prediction for energy systems with both spatial and temporal characteristics, probabilistic graphical models (PGM, including a variety of models described by conditional dependence structures, so-called graphs, including Bayesian networks and undirected/directed Markov networks, can be used to deal with dynamics systems and relational data \cite{koller2009probabilistic}), can possibly be employed as the spatiotemporal interactions are naturally suited for graph representations and can be evaluated by the associated probabilities. Bayesian networks are a type of PGM that captures causal relationships using directed edges \cite{koller2009probabilistic}, where the overall joint probability distribution of the network nodes (variables) is computed as a product of the conditional distributions (factors) defined by the nodes in the network. However, prediction problems are not straightforward for Bayesian networks, as they only encode node-based conditional probabilities, and the approximation of the joint distribution using node-based structures is often intractable \cite{sarkar2016pgm}. This is because a certain directed acyclic graphical structure may not allow for easy and exact computation of certain probabilities related to inference questions.

% [GPH: Unclear, elaborate why this makes prediction difficult].

Markov models, as a class of statistical models, have been widely applied to different domains, e.g., natural language processing and speech recognition \cite{leek1997information}. These models are shown to be efficient in identifying the probabilistic dependencies among random variables in both directed and undirected manner. Hidden Markov Models (HMMs) have been particularly successful for learning temporal dynamics of an underlying process \cite{rabiner1989tutorial}. Several modifications for HMMs have been proposed, such as integrated HMM (IHMM) \cite{beal2001infinite} which integrated several parameters to three hyper-parameters to model countably infinite hidden state sequences, integrated hierarchical HMM (IHHMM) \cite{heller2009infinite} extended HMMs to an infinite number of hierarchical levels, and \cite{wakabayashi2012forward} applied a forward-backward algorithm to reduce model complexity through the order of operations. However, Markov Models with hidden states usually rely on iterative learning algorithms that may be computationally expensive. To alleviate such issues, symbolic dynamic filtering (SDF) was proposed \cite{ray2004symbolic,rajagopalan2006symbolic} based on the concepts of symbolic dynamics and probabilistic finite state automata (PFSA). Several improvements related to coarse graining of continuous variables \cite{SSS13}, state splitting and merging techniques for PFSA \cite{mukherjee2014state}, efficient inference algorithms \cite{sarkar2013symbolic}, and hierarchical model learning \cite{akintayo2015symbolic} have been proposed over the last decade within the SDF framework. SDF has been shown to be extremely efficient for anomaly detection and fault diagnostics of various complex systems, such as gas turbine engines~\cite{SSM12}, shipboard auxiliary systems~\cite{SSV14}, nuclear power plants~\cite{JGSRE11}, coal gasification systems~\cite{CSGR08} and bridge monitoring process~\cite{LGLPS17}.

For the purpose of addressing prediction problems in disparate energy systems, this work presents a new data-driven framework (namely spatiotemporal pattern networks, or STPN) to leverage the spatiotemporal interactions of energy systems for prediction. Built on SDF, a STPN aims to capture the spatiotemporal characteristics of complex energy systems, and implement prediction at both spatial and temporal resolutions. For validation, two representative cases are proposed using the proposed approach, the first is taken from the energy supply side, wind power prediction in large-scale wind farm, and the second is from the energy demand side, energy disaggregation (also as non-intrusive load monitoring (NILM), a well-established problem that involves disaggregating the total electrical energy consumption of a household into its constituent load components without the necessity for extensive metering installations on individual household or appliance \cite{GH92,MZKR11,cominola2017hybrid}).

\textbf{Contributions}: First, a novel data-driven method for energy prediction based on the STPN framework is proposed and the concepts of interests and relevance are established. Second, two typical case studies based on wind turbine power (supply side energy) and residential building energy disaggregation (demand side energy) are performed for validating the proposed scheme. For wind turbine power prediction, the spatiotemporal characteristics between different wind turbines are identified, while the complex coupled temporal features for home energy disaggregation. A STPN-based convex programming is presented in this work in order to improve energy disaggregation performance. We also present a comparative study of energy prediction performance of the proposed technique for both cases with other state-of-the-art methods.

The remaining sections are outlined as follows. In Section~\ref{sec:Symbolic} some necessary background of SDF is presented as well as the concepts of a $D$-Markov machine. While the prediction approach based on STPN is given in Section~\ref{sec:STPN}, two typical case studies, i.e., supply side (wind turbines) and demand side (NILM), for validating the proposed framework are presented in Section~\ref{wind_turbine} and Section~\ref{NILM}, respectively. In Section~\ref{sec:con}, conclusive remarks and future research directions beyond the existing results are offered.

%For the aim of using SDF to predict wind energy we introduce an information-theoretic measure based on mutual information (MI) to yield such relationships. MI-based criteria have been widely used in feature selection [],[],[],[]. Recently Sarkar et.al [] have adopted it in sensor fusion for fault detection and classification. In our work MI between different time-series data is established such that the essential spatiotemporal features of time-series data are achieved to predict with certain probabilities. Regarding the data dependency another concept of causality is introduced. Causality has been playing an important role in Econometrics [], bioengineering [], neuroimaging [], and systems and control []. However, there are still a great amount of issues of paramount significance remained. This paper will compare the mutual information and causality in predictions between time-series data and will investigate the difference heuristically.

\section{Symbolic Dynamic Filtering and $D$-Markov Machines}\label{sec:Symbolic}
This section gives an essential background on symbolic dynamic filtering necessary to characterize the proposed prediction method. We refer interested readers to~\cite{SSS13} for more details. SDF is built upon the relevant concepts of discrete dynamic systems in which discretization and symbolization are critical steps to convert collected or observed continuous data to discrete symbol sequences. Therefore, the dynamic systems can be studied in deterministic or probabilistic settings in terms of symbolic space by using language-theoretic approaches, e.g., shift-maps and sliding block codes. The simplest approaches for partitioning are the uniform partitioning and maximum entropy, while these two methods were mainly applied to simple dynamic systems with data of less variance. The state-of-the-art partitioning or discretization approaches include symbolic false nearest neighbor partitioning (SFNNP)~\cite{PhysRevLett.91.084102}, wavelet transform~\cite{SSS13}, and Hilbert-transform-based analytic signal space partitioning (ASSP)~\cite{SR08}. Recently, a supervised partitioning scheme, i.e., maximally bijective discretization (MBD)~\cite{SSS13} has been proposed for modeling and analyzing complex dynamic systems. Unlike the other methods, MBD is able to maximally preserve the input-output relationship originating from the continuous domain after discretization in dynamical systems.

After discretization of the time-series data in continuous domain, symbolization is implemented subsequently for establishing the $D$-Markov machines. For SDF, a critical assumption is that we can approximate any symbol sequence generated by a time series data as a Markov chain of order $D$ (which is a positive integer). Therefore, such a Markov chain is called $D$-Markov machine, which is used to establish the model for each time series data due to the temporal features associated with the symbol sequence. Some relevant definitions are more formally given as follows.

\begin{definition}\label{def:DFSA}~\cite{SSV14} (DFSA) A deterministic finite state automaton (DFSA) is a 3-tuple $\mathcal{G} = (H, \mathcal{Q}, \phi)$ where:
\begin{enumerate}
\item $H$ is a set of finite size for the symbol alphabet and $H\neq\varnothing (empty\;set)$;
\item $\mathcal{Q}$ is a set of finite size for states and $\mathcal{Q}\neq\varnothing$;
\item $\phi : \mathcal{Q} \times H \rightarrow \mathcal{Q}$ is the mapping function for state transition;
%\item $q_0 \in Q$ is the start state.
\end{enumerate}
while $H^\star$ represents the collection of all finite symbol sequences from $H$ including the empty sequence $\varepsilon$.
\end{definition}

\begin{definition}\label{def:PFSA}~\cite{SSV14} (PFSA)  A probabilistic finite state automaton (PFSA) is an extension to probabilistic setting from a DFSA $\mathcal{G} = (H, \mathcal{Q}, \phi)$ as a pair $\mathcal{K}=(\mathcal{G}, F)$, i.e., the PFSA $K$ is a 4-tuple $\mathcal{K} = (H, \mathcal{Q}, \phi, F)$, where:
\begin{enumerate}
\item $H, \mathcal{Q}$, and $\phi$ have the same definitions as in Definition~\ref{def:DFSA};
\item $F : \mathcal{Q} \times H \rightarrow [0, 1]$ is defined as a symbol generation function, i.e., probability morph function which are such that $\sum_{\sigma \in H}F(q, \sigma) = 1 \ \ \forall q \in \mathcal{Q}$, where $p_{ij}$ indicates the probability of the symbol $\sigma_j \in H$ occurring with the state $q_i \in \mathcal{Q}$.
\end{enumerate}
\end{definition}

\begin{definition} \label{def:D-Markov}~\cite{SSV14} (D-Markov) A D-Markov machine is an extension of a PFSA where the previous $D$ symbols form a state as defined by:
\begin{enumerate}
\item $D$ signifies the depth of a Markov machine;
\item $\mathcal{Q}$ is a set of finite size for states with $|\mathcal{Q}| \leq |H|^D$, i.e., each state in a Markov machine is identified by some equivalence class of symbol strings whose length are $D$ with symbols in $H$;
\item $\phi : \mathcal{Q} \times H \rightarrow \mathcal{Q}$ signifies the state transition function such that if $|\mathcal{Q}| = |H|^D$, then there exist any two symbols $\alpha, \beta \in H$ and $\gamma \in H^\star$ such that $\phi(\alpha \gamma, \beta) = \gamma \beta$ and $\alpha \gamma, \gamma \beta \in \mathcal{Q}$.
\end{enumerate}
\end{definition}

\begin{remark}
Based on the Definition~\ref{def:D-Markov} it can be concluded that a D-Markov machine is naturally a stationary stochastic process $X = \cdots x_{-1} x_0 x_1 \cdots$, in which the probability of occurrence of a new symbol $x_n$ is determined by the last $D$ symbols, i.e., $P[x_n | x_{n-1} \cdots x_{n-D} \cdots x_0] = P[x_n | x_{n-1} \cdots x_{n-D}]$.
\end{remark}

We denote by $\Pi$ the state transition matrix and each entry of the matrix demonstrates the transition probability from one symbolic state to another. We give a simple example to illustrate this. Let the $k^{th}$ state of one dynamical system $A$ be $s_{k}^A$ such that the $ij^{th}$ entry, i.e., $\pi_{ij}^A$ of the matrix $\Pi^A$ indicates the probability of $s_{k+1}^A$ as $i$ given that the previous state $s_{k}^A$ was $j$, i.e.,
\begin{equation*} \label{PiA}
\pi_{ij}^A := P\left(s_{k+1}^A = i \ | \ s_{k}^A =j\right) \forall k
\end{equation*}

Moreover, one can model individual dynamical system making use of $D$-Markov machines. Because a $D$-Markov machine cannot capture the interaction dependencies for multiple systems or sub-systems in a large complex system, it has recently been extended to a x$D$-Markov machine, which was originally developed in order to obtain the internally causal dependencies among different systems or sub-systems. Different from correlation-based analysis, such a model can efficiently build up and fairly generalize the causal dependencies~\cite{C14}. The following shows the formal definition of x$D$-Markov machine.

\begin{definition}~\cite{SSV14} (xD-Markov)\label{xD-Markov}
Let $\mathcal{R}_1$ and $\mathcal{R}_2$ be the PFSAs which correspond to symbol streams $\{\textbf{x}_1\}$ and $\{\textbf{x}_2\}$ respectively. Therefore a $xD$-Markov machine is defined as a 5-tuple $\mathcal{R}_{1\rightarrow 2} := (\mathcal{Q}_1,H_1,H_2,\phi_{1},F_{12})$ such that:

\begin{enumerate}
    \item $H_1 = \{H_0, ... ,H_{|H_1|-1}\}$ represents the alphabet set of symbol sequence $\{\textbf{x}_1\}$
    \item $\mathcal{Q}_1 =\{s_1,s_2,\dots,s_{|\mathcal{Q}_1|}\}$ is the state set which corresponds to symbol sequence $\{\textbf{x}_1\}$ % where $\stateset_1$ is the set  for $\{\textbf{s}_1\}$
    \item $H_2 = \{H_0, ... ,H_{|H_2|-1}\}$ represents the alphabet set of symbol sequence $\{\textbf{x}_2\}$
    \item $\phi_{1}:Q_1 \times H_1 \rightarrow \mathcal{Q}_1$ gives the state transition mapping that maps the transition in symbol sequence $\{\textbf{x}_1\}$ from one state to another based on occurrence of a symbol in $\{\textbf{x}_1\}$
    \item $F_{12}$ is the symbol generation matrix of size $|{\mathcal{Q}_1}|\times|{H_2}|$; the $ij^{th}$ entry of $F_{12}$ denotes the probability of obtaining the symbol $\sigma_j$ of $\{\textbf{x}_2\}$ while making a transition from the state $s_i$ of $\{\textbf{x}_1\}$
\end{enumerate}
\end{definition}

Therefore, it can be observed that one can obtain the probability of a new symbol occurring after the previous $D$ symbols given for an individual symbol sequence. On the other hand, in order to know the probability of a new symbol occurring in a symbol sequence with the last $D$ symbols given in another different symbol sequence, a x$D$-Markov machine can be applied correspondingly. Equivalently speaking, given a x$D$-Markov machine, the causal dependency of one symbol sequence on another symbol sequence can be captured.

\begin{figure*}
\centering
\subfigure[]{\includegraphics[width=0.95\textwidth]{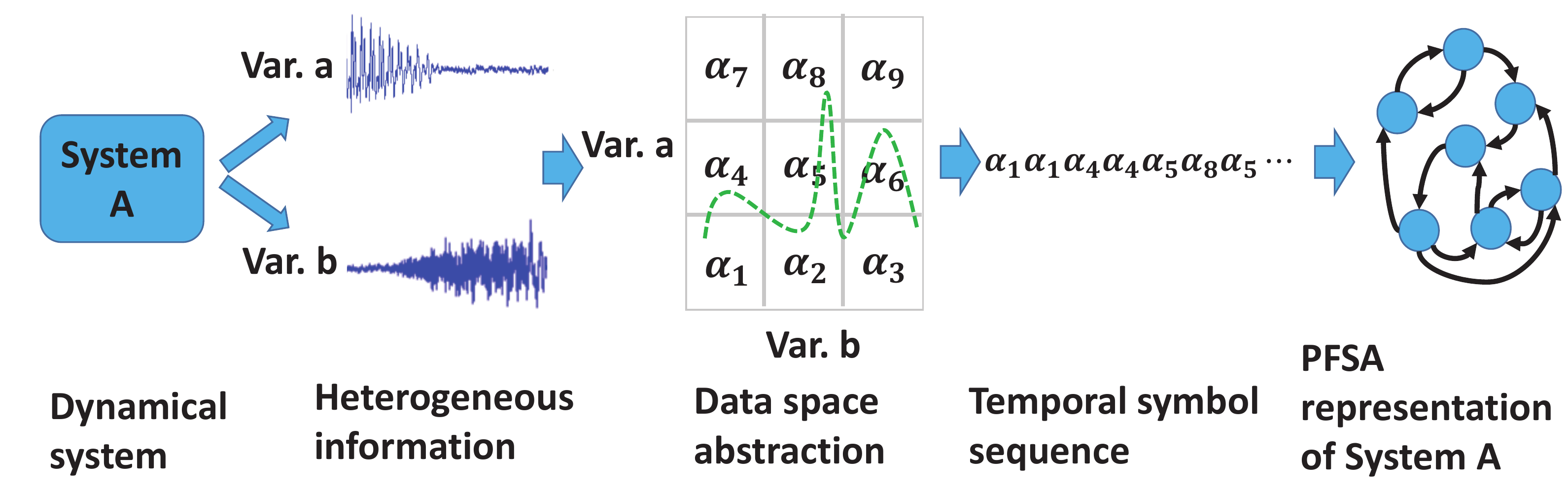}}
\subfigure[]{\includegraphics[width=0.95\textwidth]{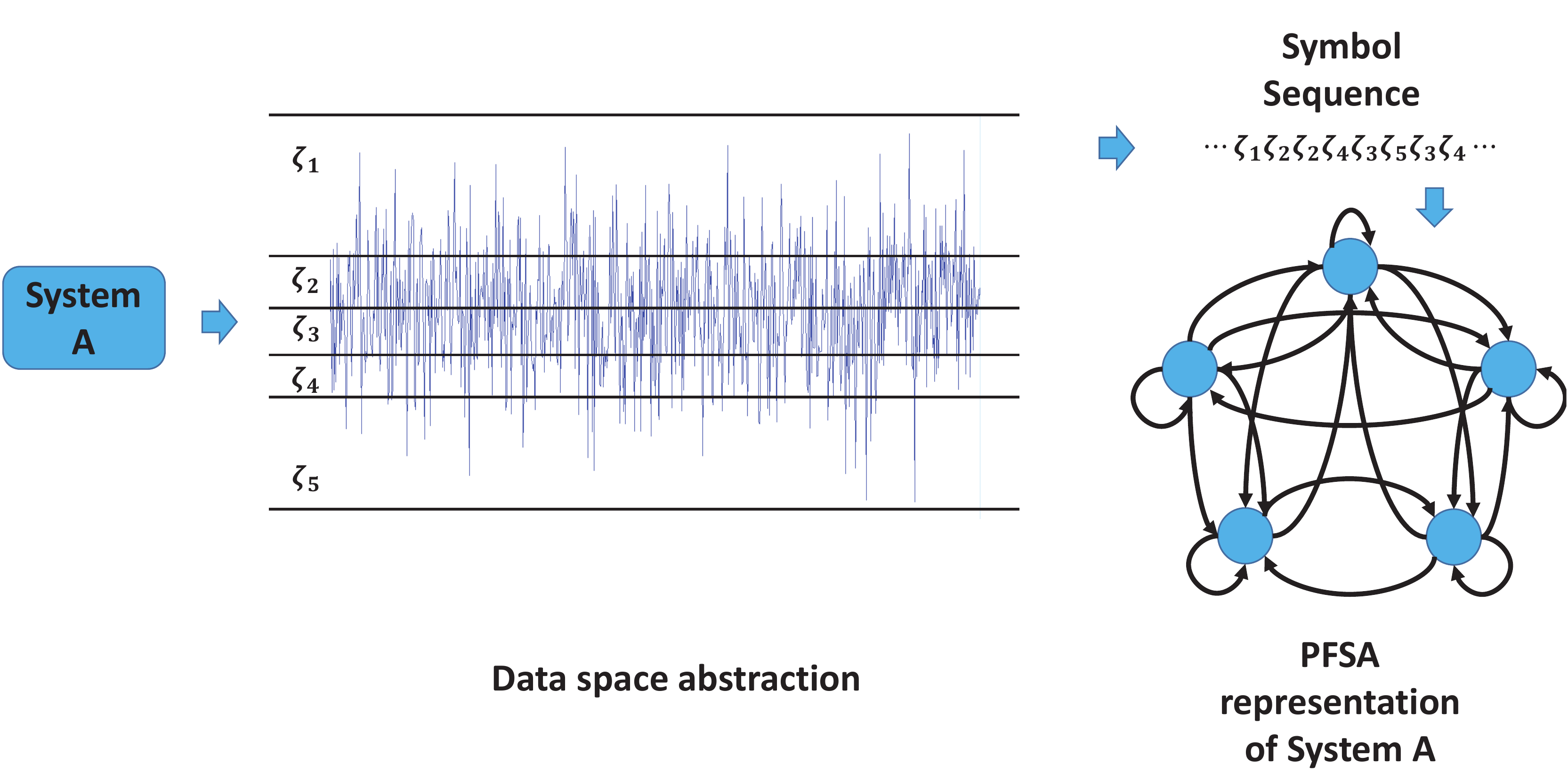}}
\caption{\textit{Illustration of generation of a PFSA using (a) maximal bijectively discretization and (b) maximum entropy partitioning for system A}.}\label{Figure1-1:1}
\end{figure*}

%\begin{figure*}
%\centering
%\includegraphics[width=0.8\textwidth]{Figures/sdf1.eps}
%\caption{\textit{Illustration of generation of a PFSA using maximal bijectively discretization for system A}}\label{Figure1-1:1}
%\end{figure*}
%\begin{figure*}
%\centering
%\includegraphics[width=0.8\textwidth]{Figures/SDF1_1.eps}
%\caption{\textit{Illustration of generation of a PFSA using maximum entropy partitioning for system A}}\label{Figure1-2:1}
%\end{figure*}

\section{Spatiotemporal Pattern Network}\label{sec:STPN}
This section mainly presents how to construct the spatiotemporal pattern network (STPN) for two dynamical systems, $A$ and $B$, based on the concepts of SDF introduced above. First we start with data partitioning/discretization and symbolization followed by the details of STPN construction.

\subsection{Discretization and Symbolization}
Suppose there are two different dynamic systems $A$ and $B$. In real-world problems, such as wind power prediction, $A$ and $B$ can represent two different wind turbines in a large wind power farm. Alternatively, in residential home energy disaggregation, $A$ and $B$ could represent HVAC system electricity consumption and that of all appliances. For each system, there are various measured variables and typically some key observations are picked to establish the model and analyze. For example, for a wind turbine, wind speed and wind power are those two key observations for power predictions. It is, however, noted that some other variables, e.g., wind direction and humidity possibly affect power such that these variables can also be taken into account. The first step to model dynamic systems in terms of symbolic dynamics is the data discretization. As mentioned above, there are many approaches that can be used; in this paper, maximally bijective discretization (MBD) is applied to the supply side dynamic systems (wind turbines) and maximum entropy partitioning is used in demand side dynamic systems (HVAC, appliances, etc.). The reason we select different methods is because of the difference of measured variables. For wind turbines, wind speed and wind power are chosen and their input-output relation in the continuous domain can be maximally maintained. However, for home energy disaggregation, the unique variable for each part of the home energy use is the energy consumption itself such that there is no input-output relation in the continuous domain.

%between these two variables for is used to maintain the functional relationship between wind speed and wind power present in the continuous domain. In this case, wind speed is discretized as an independent variable and wind power discretization depends on it to maximize bijectivity between discrete states of the two variables. The second part of the abstraction step is symbolization of continuous data using the discretization. The symbolization process converts the temporal evolution of $A$ in the multi-dimensional (2-D in this case) space into a symbol sequence. An illustration is presented in Figure~\ref{Figure1:1}: heterogeneous data for wind speed and wind power is obtained from wind turbine $A$ and using a discretization policy time-series data are converted into a temporal symbol sequence. Note, although the same discretization may be used for different wind turbines, typically they would correspond to different symbol sequence characteristics. Also, this symbolization process implicitly performs data-level information fusion with wind speed and wind power data.

\subsection{Symbolic Modeling of Dynamical Systems and Interactions}
\begin{figure}
\centering
\includegraphics[width=0.75\textwidth]{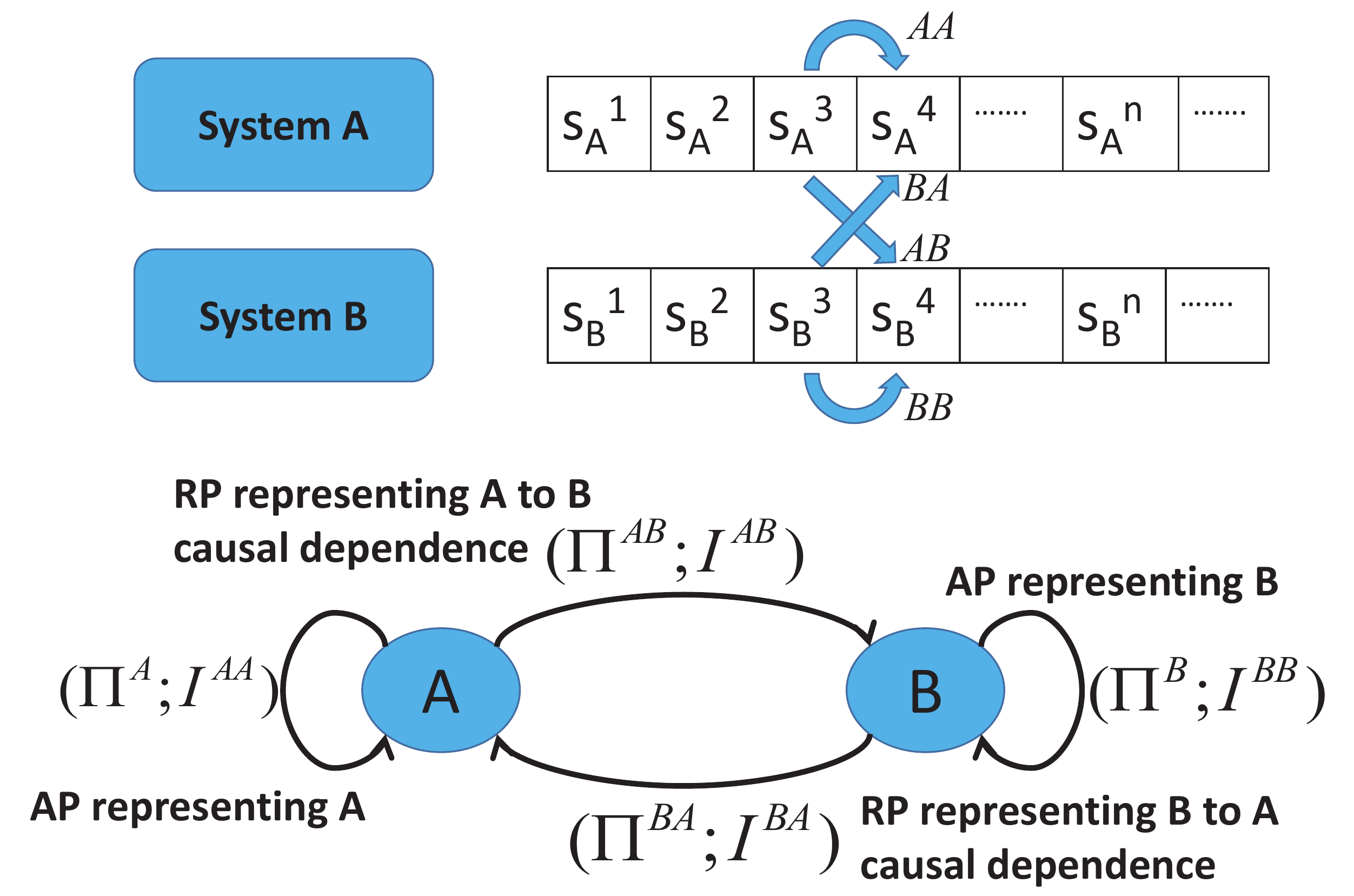}
\caption{\textit{Construction of STPN: Atomic patterns (APs) and relational patterns (RPs) formulation}.}\label{Figure2:1}
\end{figure}

Figure~\ref{Figure1-1:1} shows the symbol sequence generation in the form of PFSA using two different methods, i.e., maximally bijective discretization and maximum entropy partitioning, respectively. As discussed before it has been acknowledged that a $D$-Markov machine can be represented by a PFSA using a previous $D$ symbols to indicate one state. In this context, we take into consideration two different systems for addressing the quantification of their spatiotemporal or temporal relations. From Figure~\ref{Figure2:1}, the state transition matrices $\Pi^A$ and $\Pi^B$ show the self-relations of systems $A$ and $B$ respectively. Then the cross-state transition matrices $\Pi^{AB}$ and $\Pi^{BA}$ correspondingly represent the cause-effect relations from A to B and B to A respectively. However, it should be noted that such casual dependencies between systems $A$ and $B$ are not necessarily equivalent. For quantification of the relations in a $D$-Markov machine, a x$D$-Markov machine, atomic patterns (AP) and relational patterns (RP) were introduced in~\cite{SSV14}, which can give more details. More formally, the entries of the cross-state transition matrices $\Pi^{AB}$ and $\Pi^{BA}$ can be expressed by:
\begin{gather*} \label{PiAB}
\pi_{k\ell}^{AB} := P\left(s_{n+1}^B = \ell \ | \ s_{n}^A =k\right) \ \forall n
\end{gather*}
\begin{gather*} \label{PiBA}
\pi_{ij}^{BA} := P\left(s_{n+1}^A = j \ | \ s_{n}^B =i\right) \ \forall n
\end{gather*}
where $j,k \in Q^A$ and $i,\ell \in Q^B$.
The above relations show that a cross-state transition matrix can be constructed from symbol sequences obtained from two different dynamical systems while each entry of each matrix signifies the transition probability from one state in the first dynamical system to another state in the second dynamical system. For instance, $\pi_{ij}^{BA}$ means the transition probability from state $i$ in the system $B$ to another state $j$ in the system $A$.

Moreover, we use an information metric in order to quantify the value of the atomic and relational patterns (in this work the relational pattern is the major concern). In this context, mutual information is a metric of interest introduced to address the quantification. For example, from Figure~\ref{Figure2:1}, we denote by $I^{AA}$ and $I^{AB}$ the atomic and relational patterns respectively associated with systems A to B. Formally, the atomic pattern of system A is expressed as follows:
\begin{gather*} \label{InfA}
I^{AA} = I(s_{n+1}^A; s_n^A) = \mathcal{H}(s_{n+1}^A) - \mathcal{H}(s_{n+1}^A|s_n^A)
\end{gather*}
where
\begin{gather*} \label{HA}
\mathcal{H}(s_{n+1}^A) = -\sum_{i=1}^{\mathcal{Q}_A} P(s_{n+1}^A = i)\log_2 P(s_{n+1}^A = i)
\end{gather*}
\begin{gather*} \label{HAA}
\mathcal{H}(s_{n+1}^A|s_{n}^A) = -\sum_{i=1}^{\mathcal{Q}_A} P(s_{n}^A = i)\mathcal{H}(s_{n+1}^A|s_{n}^A = i)
\end{gather*}
\begin{gather*} \label{HAAcond}
\mathcal{H}(s_{n+1}^A|s_{n}^A = i) = -\sum_{i=1}^{\mathcal{Q}_A} P(s_{n+1}^A = l|s_n^A = i)\cdot \nonumber \\
\log_2 P(s_{n+1}^A = l|s_n^A = i)
\end{gather*}
Therefore, based on the quantity $I^{AA}$ (defined using different entropy $\mathcal{H}$ values as presented above), the temporal self-prediction capability of the system A can be correspondingly identified.

On the other hand, the mutual information for the relational pattern involved in systems A and B can be described as:
\begin{gather*} \label{InfAB}
I^{AB} = I(s_{n+1}^B; s_n^A) = \mathcal{H}(s_{n+1}^B) - \mathcal{H}(s_{n+1}^B|s_n^A)
\end{gather*}
where
\begin{gather*} \label{HBA}
\mathcal{H}(s_{n+1}^B|s_{n}^A) = -\sum_{i=1}^{\mathcal{Q}_A} P(s_{n}^A = i)\mathcal{H}(s_{n+1}^B|s_{n}^A = i)
\end{gather*}
\begin{gather*} \label{HBAcond}
\mathcal{H}(s_{n+1}^B|s_{n}^A = i) = -\sum_{i=1}^{\mathcal{Q}_B} P(s_{n+1}^B = l|s_n^A = i)\cdot \nonumber \\
\log_2 P(s_{n+1}^B = l|s_n^A = i)
\end{gather*}
Hence, the quantity of $I^{AB}$ identifies system A's capability of predicting system B's outputs and vice versa for $I^{BA}$. Furthermore, based on the mutual information, patterns can be assigned with weights such that some patterns with low mutual information may be rejected for simplifying the model.
%We denote by the total information gain $I_{G}^{tot}$ the sum of mutual information for all patterns. Then, we have
%\begin{equation*}
%I_{G}^{tot} = \sum_{(A,B)\in \mathcal{S}\times \mathcal{S}} I^{AB}
%\end{equation*}
%where $\mathcal{S}$ is set of all wind turbines. As the patterns are selected when using mutual information then the insignificant patterns would be eliminated from the set $\mathcal{S}\times \mathcal{S}$ of all patterns.
Interested readers can find more details in~\cite{SSV14}.

%As the discussion above, the STPN for wind turbine interactions in a wind farm is established upon SDF and mutual information based information-theoretical measure is first time applied in the investigation of interactions between multiple wind turbines to our best knowledge. The STPN provides an effective tool for learning the spatiotemporal interactions among individual wind turbines and realizes the conversion not only from physical to symbolic domains but also from symbolic to physical domains. This means via STPN the relations between multiple wind turbines can be quantified in the symbolic domain first, and then given the mutual information obtained, using the information of one wind turbine is able to predict that of another wind turbine.

Based on the above analysis, it has been shown that the proposed STPN in this paper can be an effective tool for capturing the spatiotemporal interactions between different dynamic systems. For validating such a data-driven method this paper offers two case studies in terms of supply side dynamic systems (i.e., wind turbines in a wind farm) and demand side dynamic systems (i.e., home electric energy disaggregation) to demonstrate the efficacy and effectiveness. The prediction process can be described as follows: Given a training data set in the continuous domain, we use partitioning methods to discretize and symbolize the data for running the x$D$-Markov machine. The probability transition matrices are obtained for predictions in symbolic or continuous domains. For the symbolic prediction, we find out the most likely symbol sequence for system $A$ given another symbol sequence of system $B$ running the x$D$-Markov model numerous times. While in continuous domain the prediction can be acquired based on the symbolic prediction using expectation as follows:

\begin{equation}\label{prediction_STPN}
W(k) = \sum_{j=1}^{m} Pr_k(j)W(E|j)
\end{equation}

where, $W(k)$ represents the expectation of energy at the $k^{th}$ instant, $Pr_k(j)$ signifies the probability of $j^{th}$ symbol occurring at the $k^{th}$ instant after running numerous simulations of Monte Carlo Markov Chain, $W(E|j)$ indicates the expectation of energy for the discrete bin labelled by symbol $j$ (suppose that in that bin there are $m$ discrete symbols).

The pseudocode of energy prediction based on STPN is as follows.

\begin{algorithm}[H]\label{STPN}
    \caption{Energy Prediction based on STPN}
    \SetKwInOut{Input}{Input}
    \SetKwInOut{Output}{Output}

    \Input{Training data sets of systems $i$, $C'_i$ ($i$ represents any system), depth of $D$}
    \Output{Predicted results $\hat{C}_i$}
    \text{Discretize and symbolize the continuous data $C'_i$ to $s_i$}\;
    \text{Calculate state transition matrices and mutual information by $s_i$}\;
    \text{Calculate the expected value of energy in the discrete bin}\;
    \text{Use Eqn.~\ref{prediction_STPN} to calculate the prediction results $\hat{C}_i$}\;
\end{algorithm}

\begin{figure}
\centering
\includegraphics[width=0.8\textwidth]{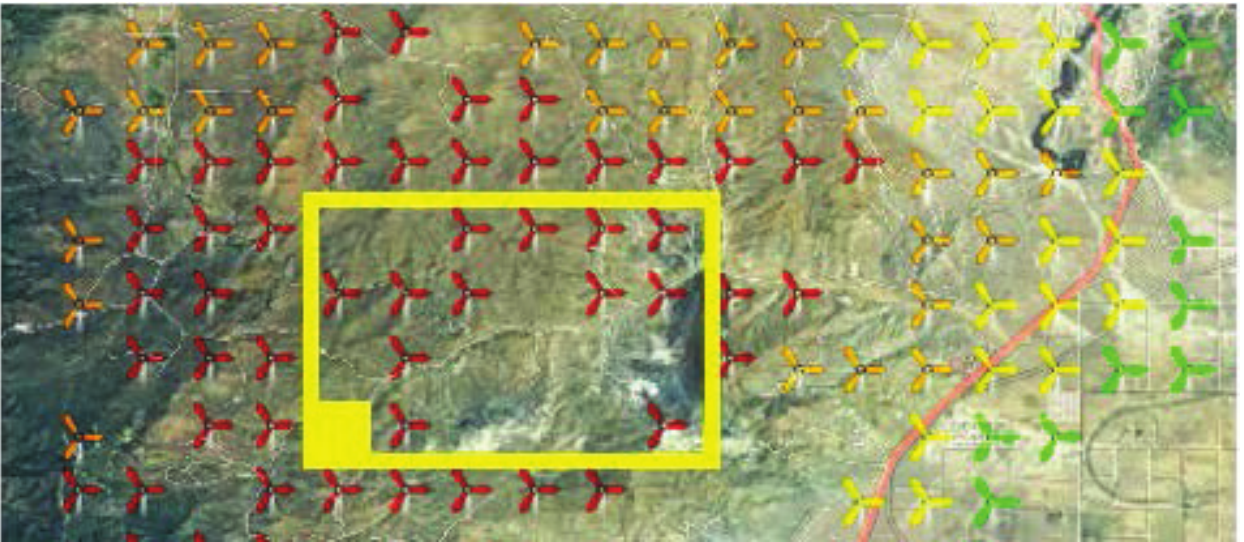}
\caption{\textit{Geographical information of wind turbines under analysis which are located in California, between 35.28-35.33n and 118.09-118.17w}}\label{Figure3:1}
\end{figure}

\begin{figure}
\centering
\includegraphics[width=0.8\textwidth]{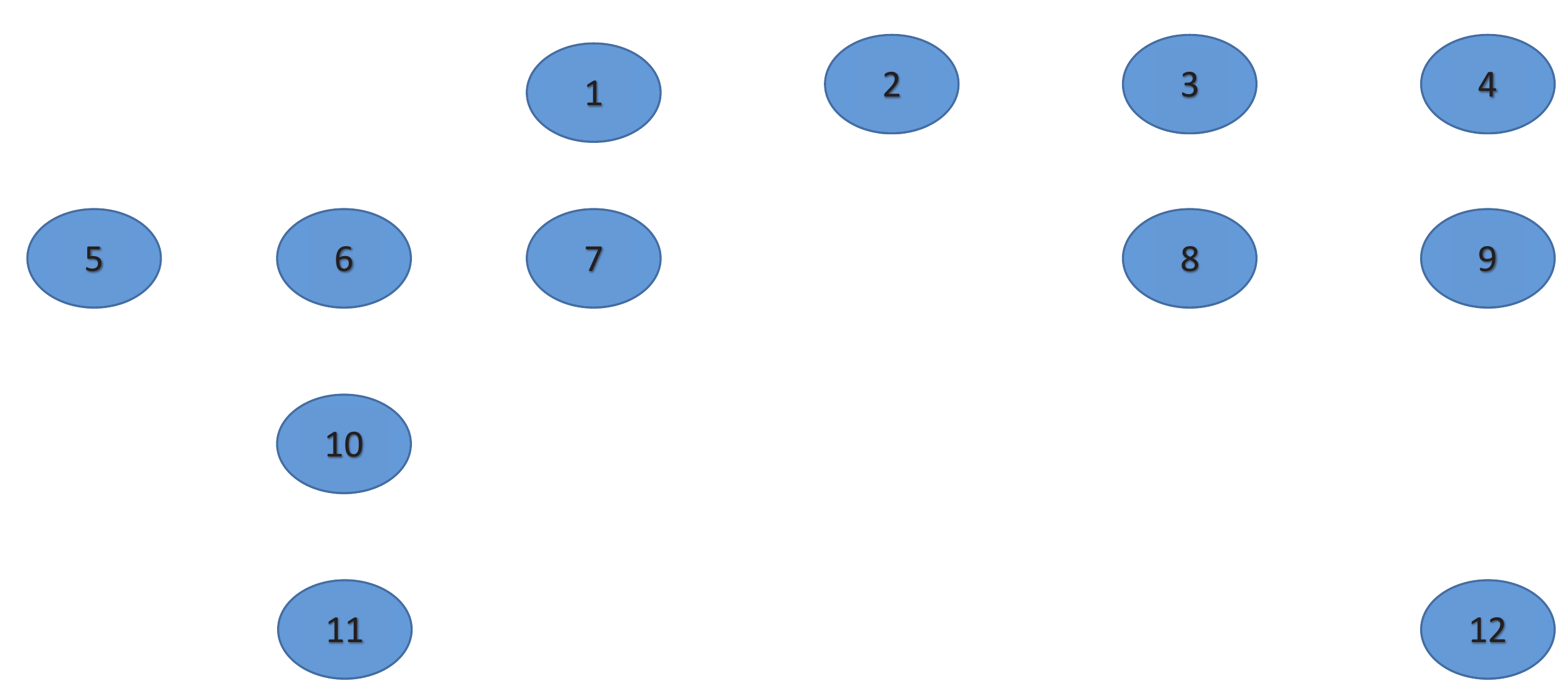}
\caption{\textit{Representation of STPN for 12 wind turbines}}\label{Figure4:1}
\end{figure}

\section{Supply Side: Wind Turbines}\label{wind_turbine}
\subsection{Geographical information}
In this subsection, a case study based on the supply side energy systems, i.e., wind turbines, is used for validating the data-driven method proposed in this work. The STPN framework is used in a wind turbine network in order to capture the causal dependencies among different wind turbines that can be regarded as sub-systems of a wind farm. This paper uses the 2006 Western Wind Integration data set obtained from NREL~\cite{NREL06} to uncover causal dependencies which are vitally important to individual wind turbine power prediction in a mutual turbine-turbine setting. For establishing the STPN, twelve wind turbines (located in California) that have capacity factors in excess of 40\% are chosen; their IDs can be identified as: 4494, 4495, 4496, 4497, 4423, 4424, 4425, 4426, 4427, 4361, 4313 and 4314 (labeled by 1-12) in this context and the capacity factors are between 41\% and 45\% approximately. For completeness, the geographical information of the wind turbines is also provided. The annual average wind velocity in the area where the considered turbines are located is around 9 $m/s$, with an elevation from 1019 to 1207 m.

\begin{figure}
\centering
\includegraphics[width=0.8\textwidth]{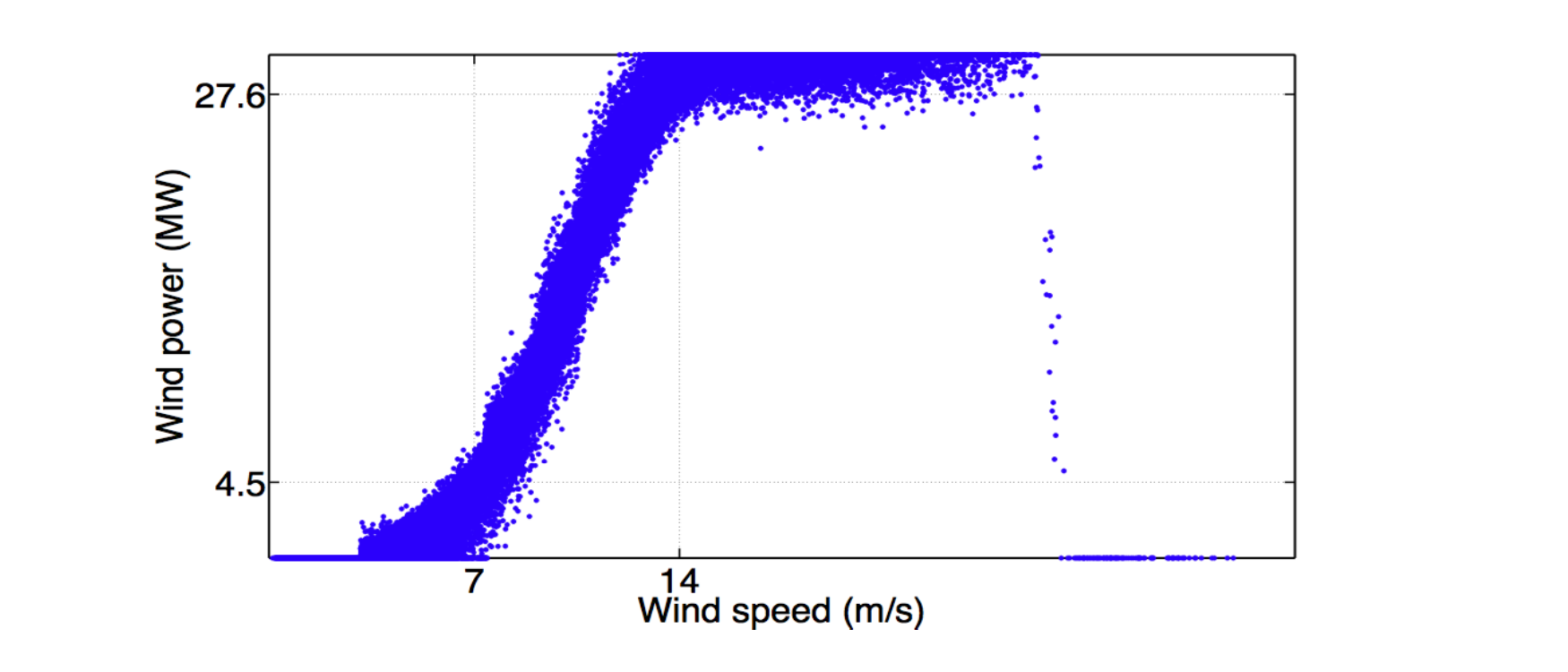}
\caption{\textit{Discretization of a typical wind turbine systems using MBD}}\label{Figure5:1}
\end{figure}

\begin{figure}
\centering
\includegraphics[width=0.8\textwidth]{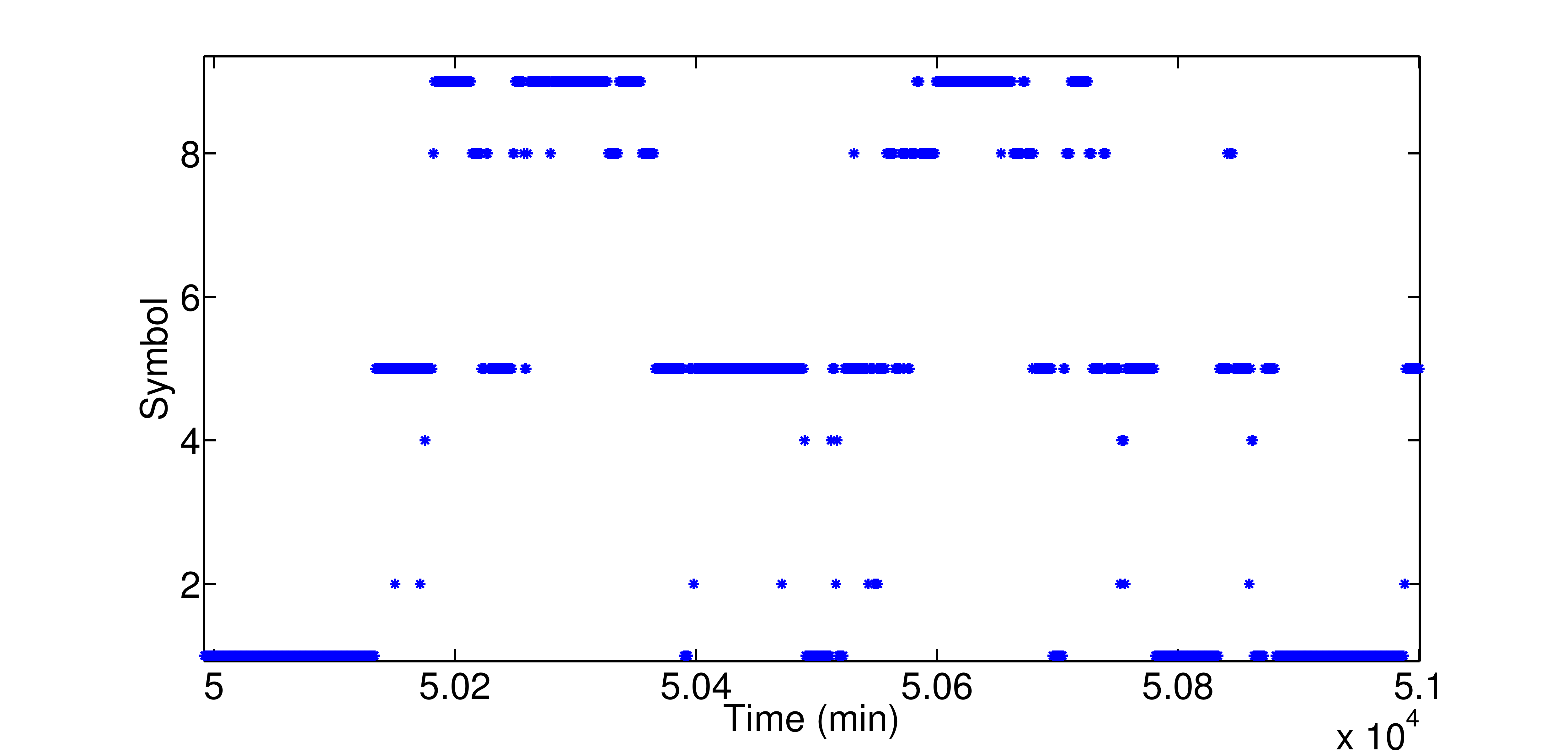}
\caption{\textit{Symbol sequence plot for a typical wind turbine}}\label{Figure6:1}
\end{figure}

As shown in Figure~\ref{Figure3:1}, twelve wind turbines are distributed in various locations, which can be identified as nodes in the STPN represented by Figure~\ref{Figure4:1}. From Figure~\ref{Figure5:1} the relation between wind speed and wind power can be observed and the other wind turbines as well have the same pattern. The input-output relation involving a wind turbine is significant such that MBD enables the maximum preservation of their correspondence in the symbolic domain. The spatiotemporal patterns between different wind turbines and the very relational patterns between them can be found from the symbol sequences. Figure~\ref{Figure6:1} shows an instance of symbol sequence for a wind turbine and it can be observed that most of the symbols are 1, 5, 8 and 9.

%Figure~\ref{Figure3:1} provides the geographical information of twelve wind turbines distributed in different locations. They are abstracted as nodes of the spatiotemporal pattern network (STPN) as shown in Figure~\ref{Figure4:1} for visualization simplicity. Figure~\ref{Figure5:1} shows the scatter plot with wind power output and wind speed of one wind turbine (other wind turbines also follow nearly the same pattern). The figure further shows the MBD where the partitioning of the wind power axis can be found given the partitioning of the wind speed axis such that their correspondence in the continuous domain is maximally preserved in the symbolic domain. In this context, the discretization of the power curve is tailored to create the symbols in order to find out the spatial and temporal characteristics of wind turbines and interaction patterns between them in the symbolic domain for wind power prediction task. There are nine symbols generated based on the data partitioning. Figure~\ref{Figure6:1} shows an example symbol sequence for a typical wind turbine. As it is evident from the discretization plot, most of the symbols are 1, 5, 8 and 9.

\begin{figure}
\centering
\includegraphics[width=0.8\textwidth]{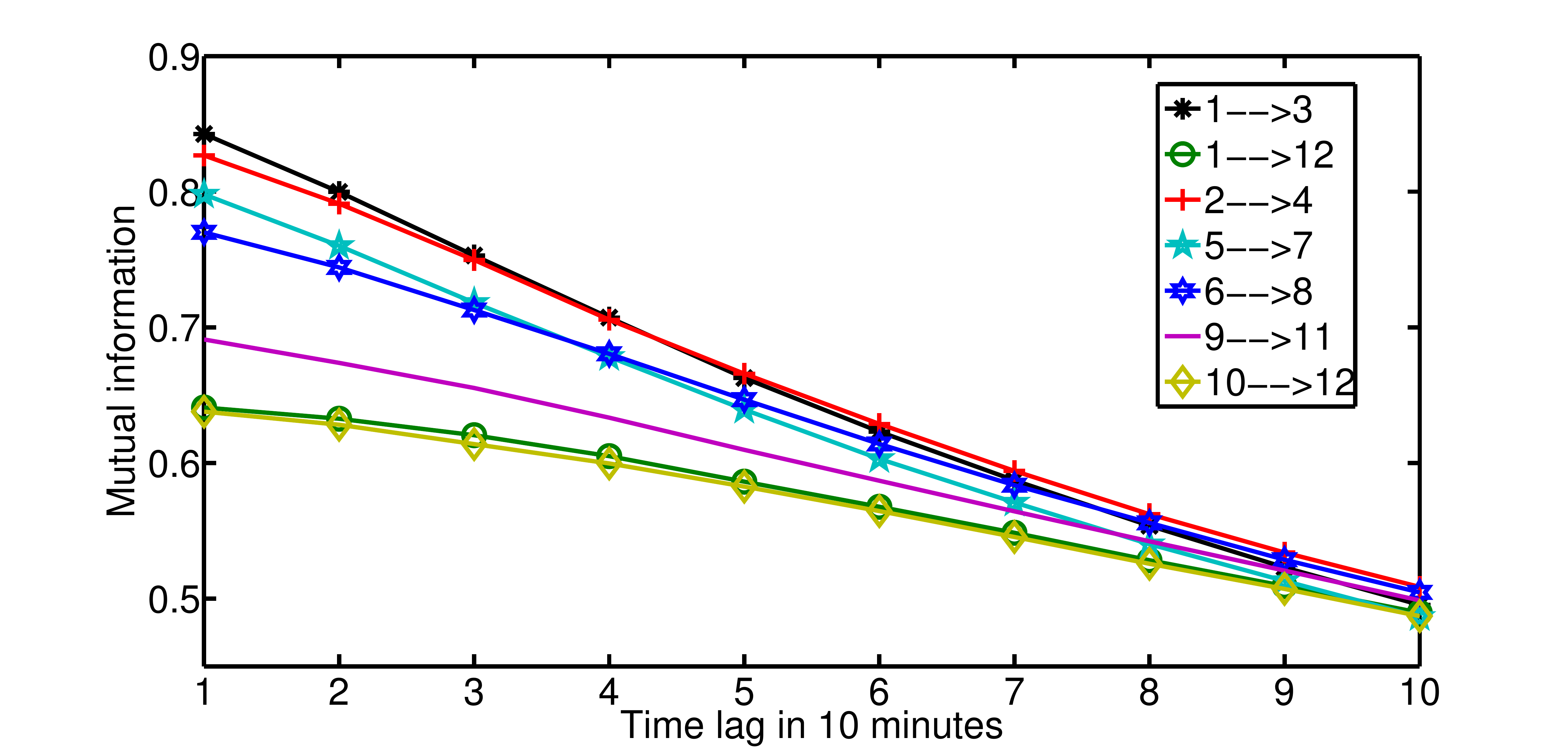}
\caption{\textit{Mutual information of relational patterns for selected pairs of wind turbines}.}\label{Figure7:1}
\end{figure}

\subsection{Results and Discussion}
The mutual information of RP between a pair of wind turbines is first to be investigate according to the state transition matrices generated by x$D$-Markov machines. We set the depth as 1 for simplicity and one can increase the parameter. Therefore, it can be immediately known that the current state of one selected wind turbine depends only on the last state of another selected wind turbine. The effect of time lag on the mutual information between wind turbines is studied for addressing the temporal characteristics. The results in Figure~\ref{Figure7:1} show that as the time lag increases the mutual information decreases correspondingly. Thus in this work one can maximize the causal dependencies between any two different wind turbines at time lag 1.

%The mutual information based causality metric (for the relational pattern) is calculated using the state transition matrices of the x$D$-Markov machines. As the depth of Markov machine is fixed as 1 in this paper, the current symbol for one wind turbine is only dependent on the past one symbol of another wind turbine. With this setup, we first investigate the effect of time lag on the causal impact between two turbines. As shown in Figure~\ref{Figure7:1}, mutual information between any pair of wind turbines decreases along with increment of time lag from 1 to 10. This shows that the choice of depth to be 1 maximizes the causality at time lag 1 for every case.

\begin{figure*}
\centering
\includegraphics[width=0.8\textwidth]{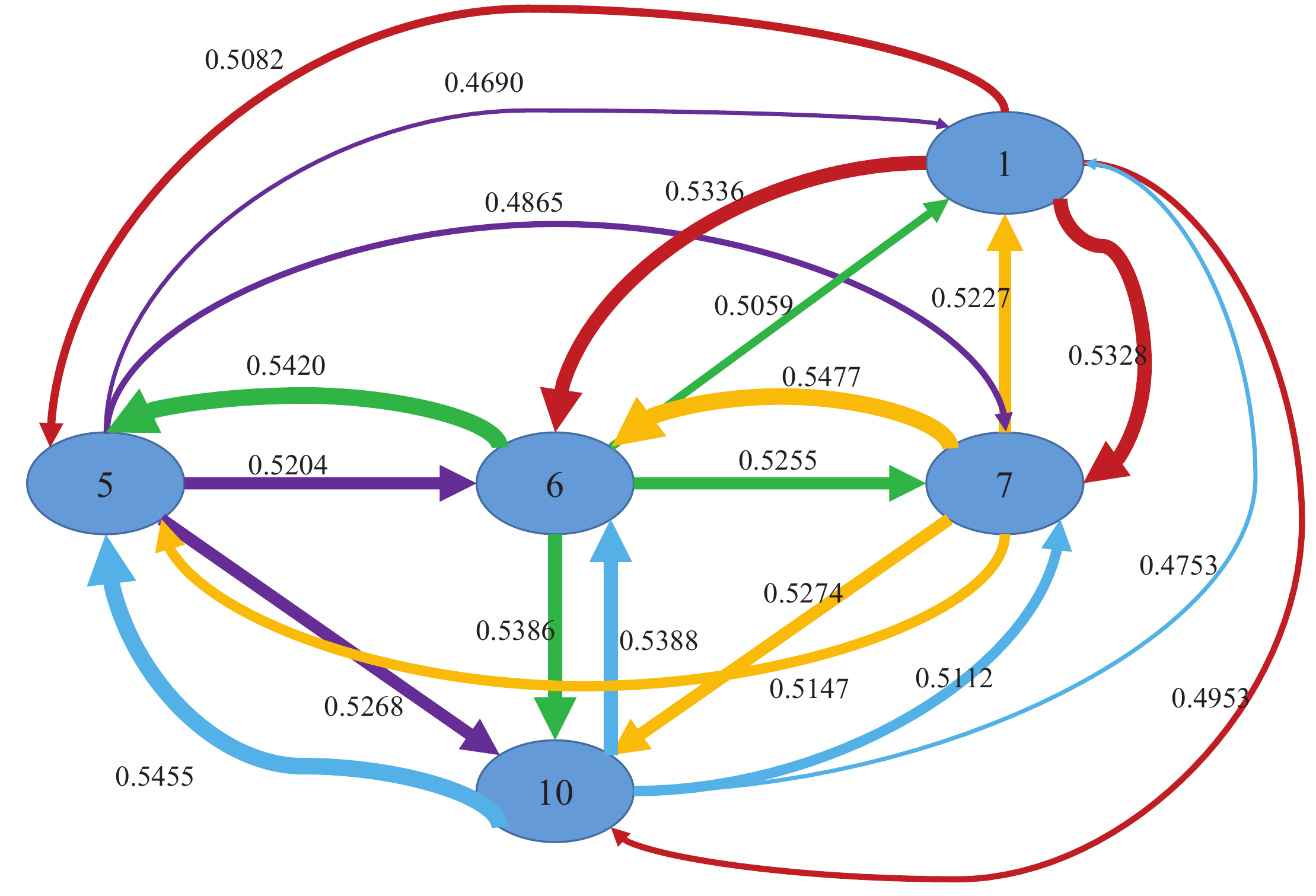}
\caption{\textit{Spatiotemporal pattern network for the group of wind turbines}}\label{Figure8:1}
\end{figure*}

\begin{figure}
\centering
\includegraphics[width=0.8\textwidth]{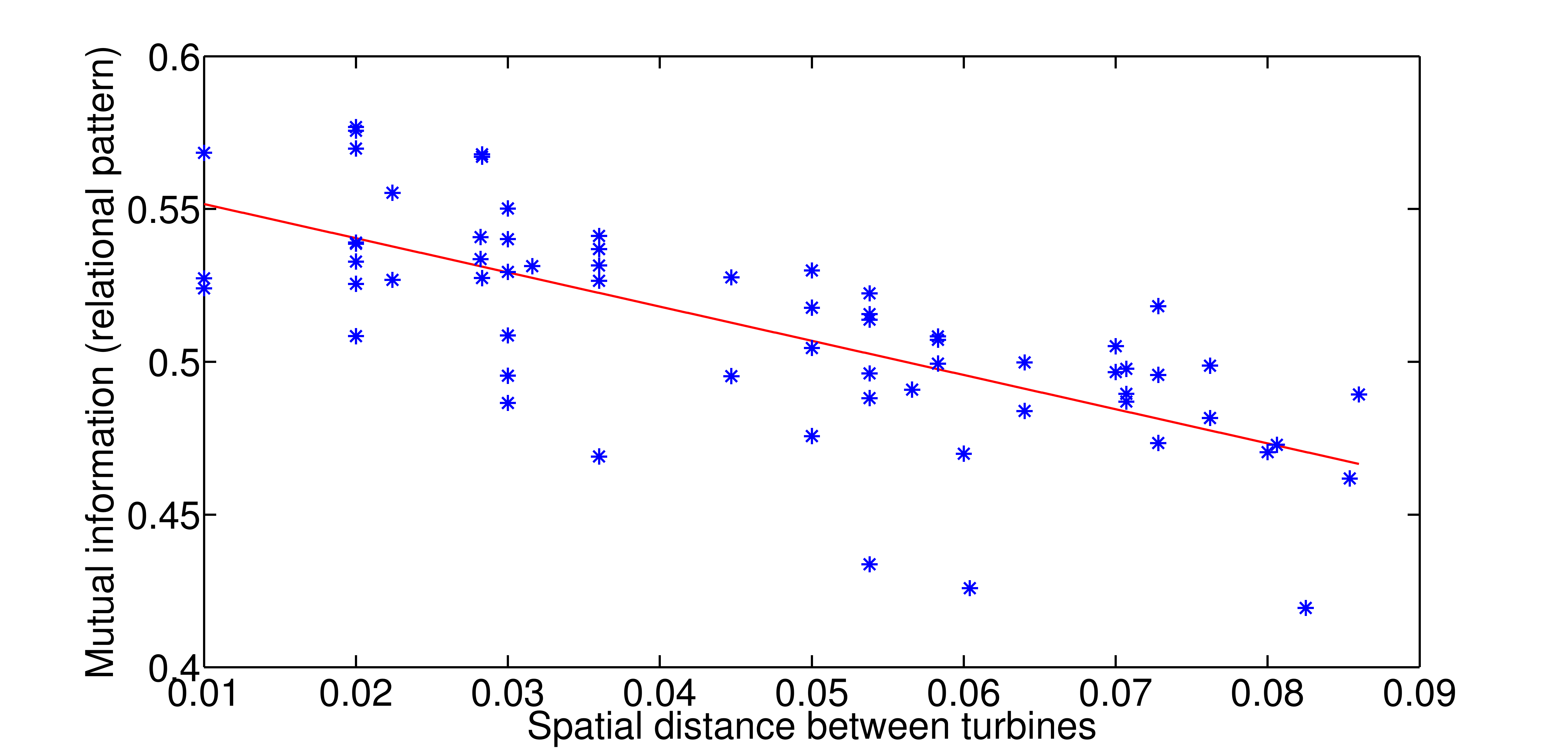}
\caption{\textit{A monotonically decreasing relationship for different pairs of wind turbines when spatial distances increase }}\label{Figure9:1}
\end{figure}

The spatial characteristics between two different wind turbines is also another critical factor in STPN. Wind turbines labeled by 5, 6, 7, 1, and 10 are chosen for the purpose of such an analysis. Figure~\ref{Figure8:1} shows that the causal dependency between any two wind turbines reduces with the increment of geographical (spatial) distance between them along any direction. Figure~\ref{Figure9:1} also illustrates that the metric based on mutual information for a pair of wind turbines with the  Euclidean distance between them exhibits a generally decreasing trend. Consequently, in summary, based on both of these observations made, applying the metric based on mutual information is an effective technique to capture the spatial and temporal patterns in wind turbine systems.

%Next we investigate the effect of spatial distance on the causal dependency between turbines. We choose wind turbines 5, 6, 7, 1 and 10 for this purpose. The results in Figure~\ref{Figure8:1} show that with increase in spatial distance between wind turbines (along any direction - latitude or longitude), causality quantified by mutual information of the relational pattern decreases. Figure~\ref{Figure9:1} shows the general decreasing trend of mutual information based causality metric and the Euclidean distance between pairs of wind turbines. Based on the above two observations, it is evident that the mutual information based causality metric is able to capture both temporal and spatial characteristics.

\begin{figure}
\centering
\includegraphics[width=0.8\textwidth]{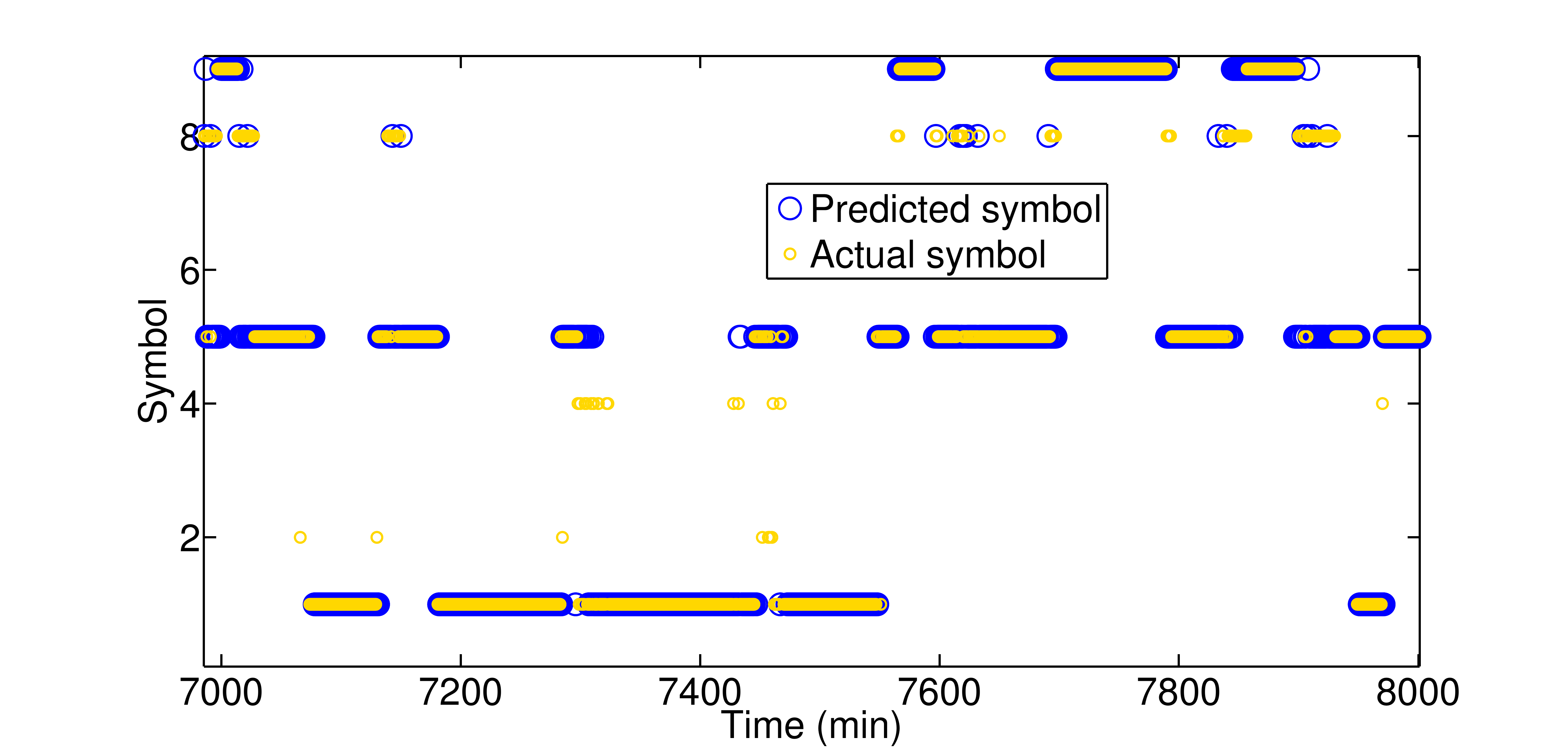}
\caption{\textit{Symbolic prediction of wind turbine 5 behavior with the observation of wind turbine 6}}\label{Figure10:1}
\end{figure}

\begin{figure}
\centering
\includegraphics[width=0.8\textwidth]{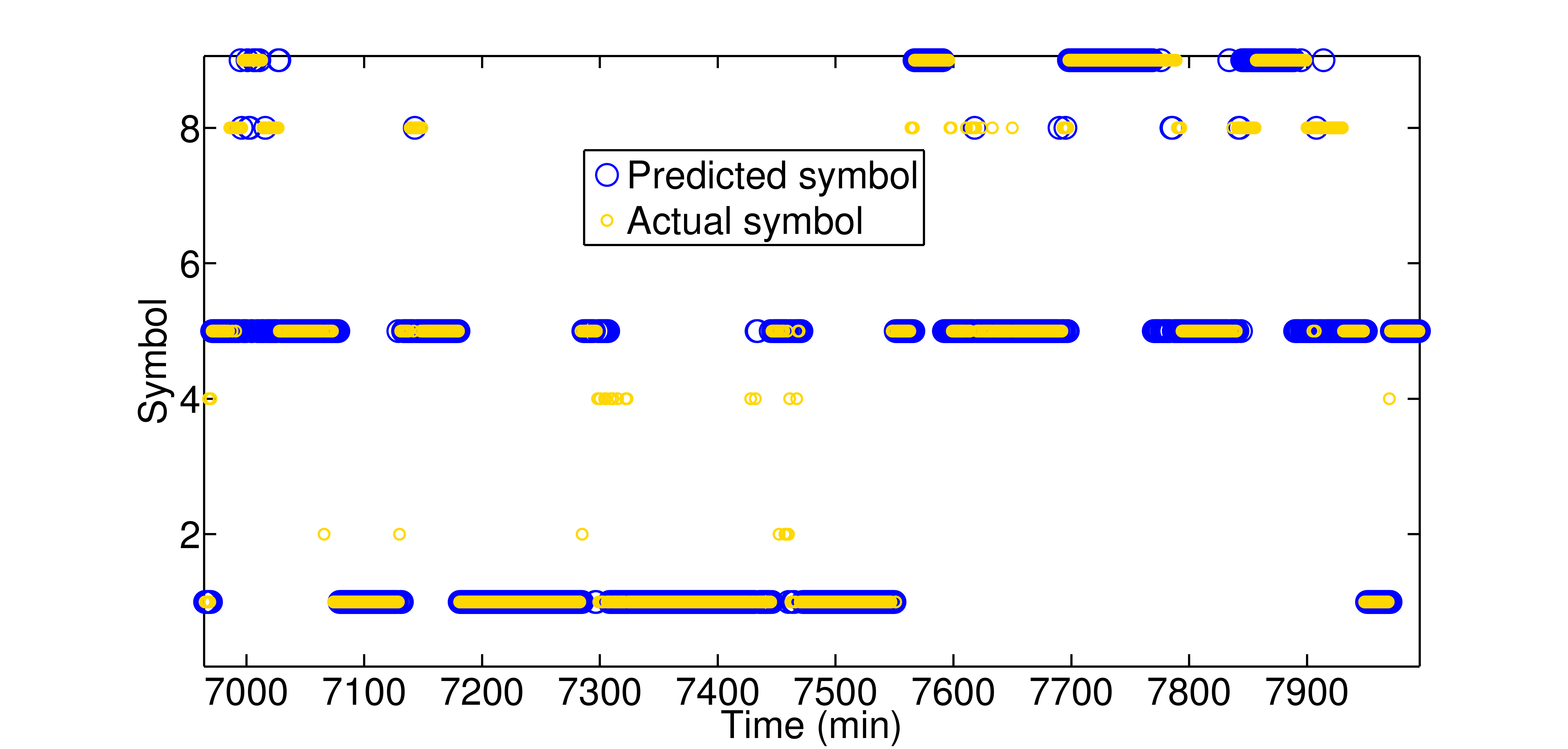}
\caption{\textit{Symbolic prediction of wind turbine 5 behavior with the observation of wind turbine 7}}\label{Figure11:1}
\end{figure}

Next, we evaluate the effectiveness of the STPN in revealing causal dependencies through wind power prediction. The symbolic and continuous prediction of one wind turbine power is based on the observed symbol sequence emerging from another turbine. According to the procedure of energy prediction described above, Figure~\ref{Figure10:1} and Figure~\ref{Figure11:1} show the symbol prediction results in which the predicted symbol sequences emerging from the wind turbine 5 under the observations of wind turbines 6 and 7 respectively are compared to the true symbol sequences emerging from the wind turbine 5. It is noted that the model is trained by the data from the first half-year of 2006 while tested by the second half-year data. From those two plots it can be observed that for most of time the proposed x$D$-Markov machines have a strong prediction capability, while some errors may come from the transient symbols. Moreover from observation it can be found that the prediction by wind turbine 6 is slightly better than that by wind turbine 7 as implied by mutual information.

\begin{figure}
\centering
\includegraphics[width=0.8\textwidth]{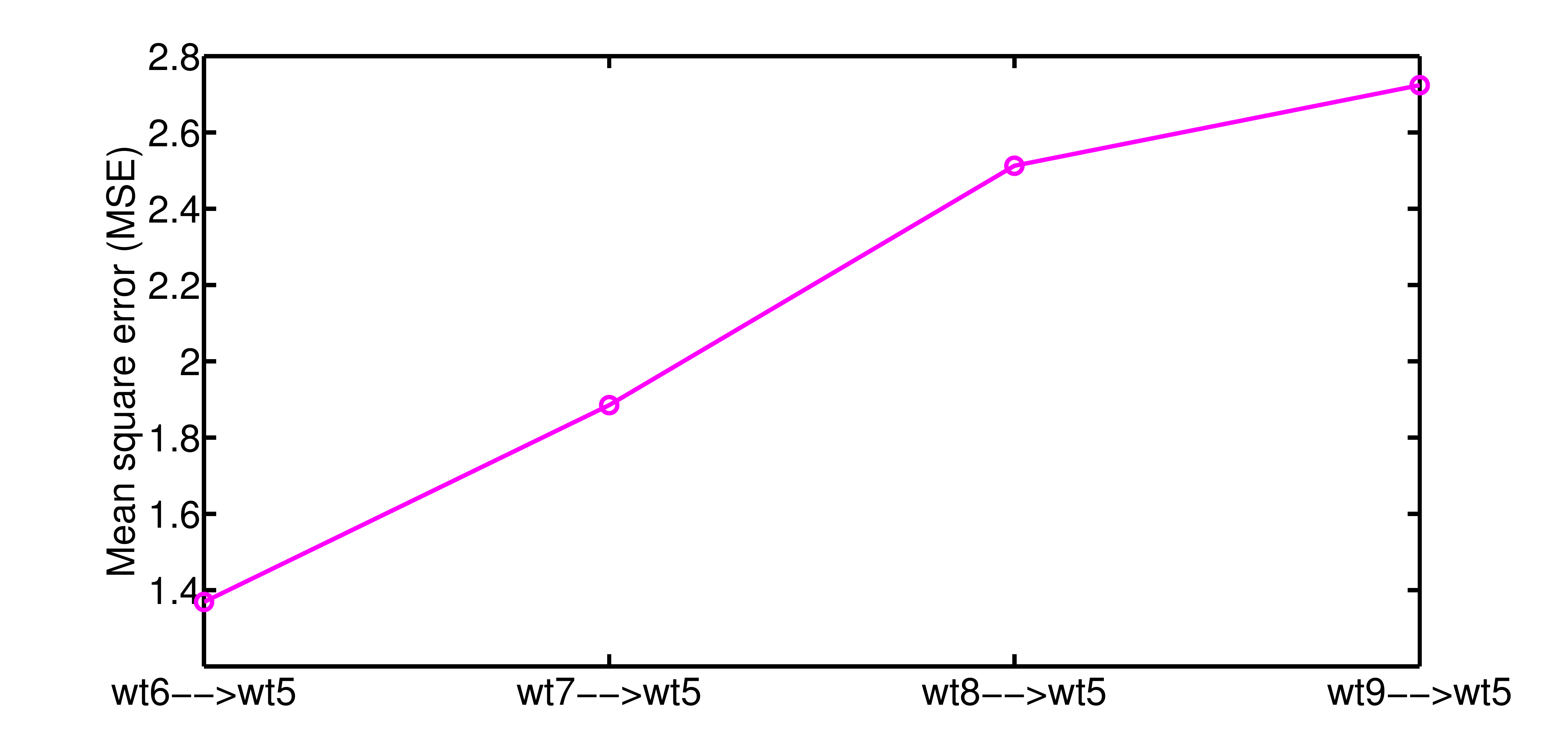}
\caption{\textit{MSEs of prediction of wind turbine 5 power using observation from other turbines: As geographical (spatial) distance increases, MSE increases}}\label{Figure12:1}
\end{figure}

\begin{figure}
\centering
\includegraphics[width=0.8\textwidth]{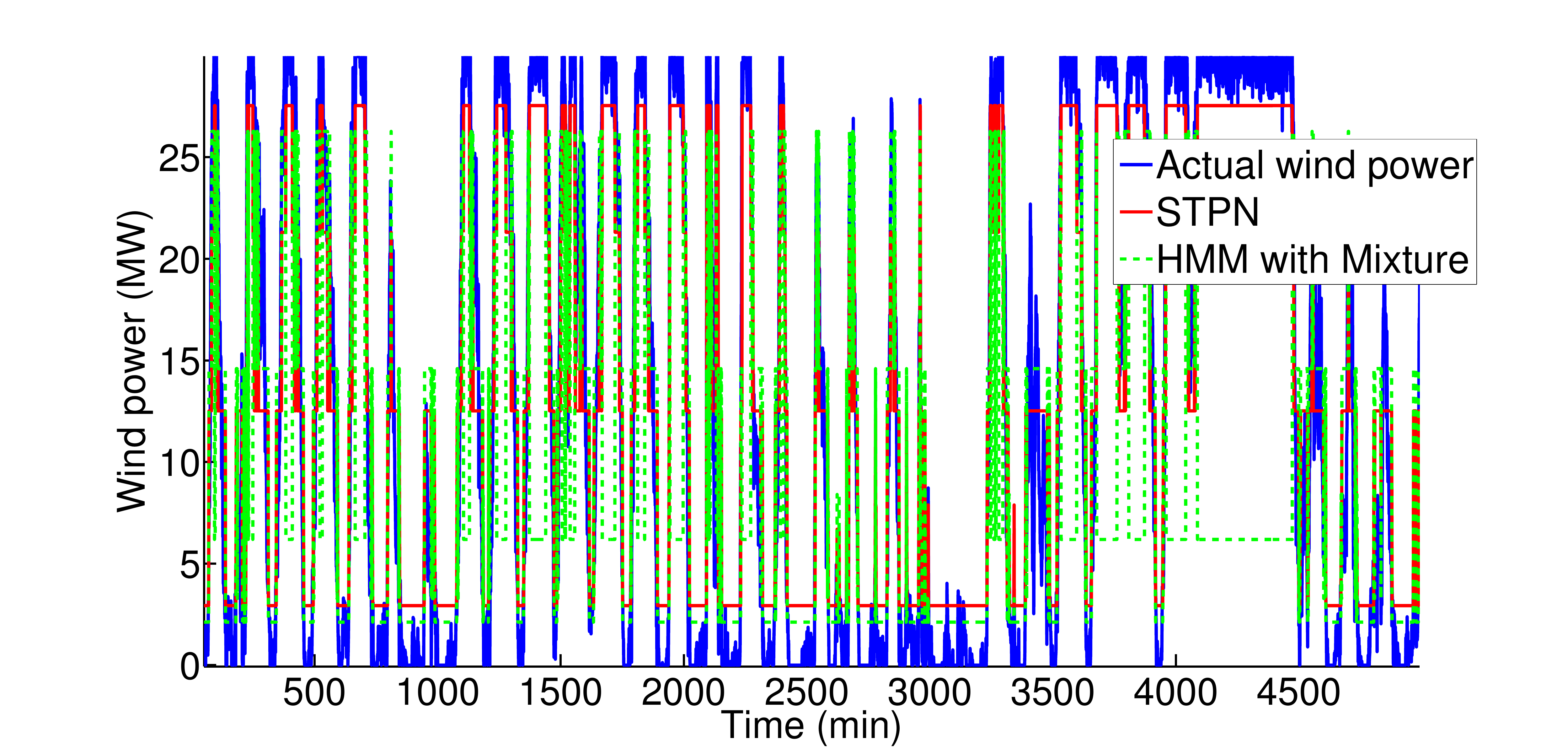}
\caption{\textit{Wind power prediction for wind turbine 5 under the observation of symbol sequence of wind turbine 6 using STPN and HMM with mixture}}\label{Figure13:1}
\end{figure}

Figure~\ref{Figure12:1} shows that the mean square error (MSE) is a function of spatial distance between any pair of wind turbines using wind turbines 5, 6, 7, 8, and 9 and it displays a monotonically increasing trend. The prediction capacity in terms of symbols using the proposed STPN has been shown. An example of energy prediction for wind turbine 5 in the continuous domain with the observation of symbol sequence for wind turbine 6 is shown here to validate the energy prediction method. The plot of Figure~\ref{Figure13:1} shows that the major trend in the actual data can be caught quite well and accurately for the continuous domain prediction as the partitioning method MBD is effective in preserving the input-output relation. However, a finer discretization may improve the prediction result in the continuous domain even though that requires a larger amount of data and increases the computational complexity correspondingly.

In order to evaluate the proposed scheme in wind power prediction, in this work we compare the performing capabilities of the STPN framework and a quite popular approach, namely, the Hidden Markov Model (HMM) with mixture which is adapted from HMM to deal with multiple variables. A toolbox compatible with MATLAB~\cite{murphy2013hidden} is applied in this context. The results in Figure~\ref{Figure13:1} have shown that the proposed prediction method based on STPN framework outperforms the HMM with mixture under visual inspection. Quantitatively, while the MSE for predicted power using HMM with mixture is $99.8842$, the MSE for predicted power using the proposed algorithm is $18.9521$. Therefore, it can be concluded that the STPN scheme in which causal dependencies between different wind turbines are captured is a quite effective technique in wind power prediction.

\section{Demand Side: Non-Intrusive Load Monitoring}\label{NILM}
This subsection presents a second case study based on demand side energy systems; in particular, non-intrusive load monitoring (NILM) of electrical demand with the purpose of identifying electric load components for residential homes. As described in the above section, the STPN framework is used as well for electric load component disaggregation. In order to best identify the disaggregated energy usage corresponding to each electric energy consuming component from the total energy consumption, convex programming is applied here. This is necessary because for NILM there is no clear input-output relation with the result that--even though the STPN is used in this case study--the results obtained may not be optimal. Here, optimal means that the summation of all load components of residential home energy consumer adds up to the whole building electricity use. Therefore, with the prediction results by STPN, a convex programming based modification is introduced to achieve said optimal disaggregation.

\subsection{Problem Description}
For this case, the data set used for energy disaggregation is based on the Building America 2010 data set available from NREL~\cite{hendron2010building}. The data is for the hot and dry location of Bakersfield, California with ample of heating, ventilation, and air-conditioning (HVAC) in the summer and includes the whole building electric (WBE), which is the sum of HVAC, lights, appliances (APPL), and miscellaneous electric loads (MELS). The goal here is to apply the measured WBE time series to predict HVAC, LIGHTS, APPL, and MELS, respectively. It is noted that WBE is the only known variable and for each part of prediction one month data is adopted where the first three week data is used for training the model, while the fourth week for testing.

\textbf{Convex Programming}: Before stating the prediction results, the convex programming problem setup is formulated for completeness. Suppose that the results obtained by STPN framework are group truth for each part except WBE. Thus the optimization problem can be expressed by
\begin{equation}\label{convex_programming}
\begin{aligned}
&\text{minimize}_{C_i, i = 1,2,3,4}J:=\sum_{i=1}^{4}\|C_i-\hat{C}_i\|^2_2\\
&\text{s.t.}\sum_{i=1}^{4}C_i = S; C_i\in\mathbb{R}^n_{\geq 0}
\end{aligned}
\end{equation}
where $C_i$ represent the decision variables to be determined, $\hat{C}_i$ signify the prediction results obtained from STPN, $S$ is the known values of WBE, $\|C_i-\hat{C}_i\|_2$ is the Euclidean norm between $C_i$ and $\hat{C}_i$.

The pseudocode of energy prediction based on STPN framework and convex programming is shown as follows. We use STPN+convex programming for reference of the combination of the STPN framework and convex programming technique throughout the rest of analysis.

\begin{algorithm}[H]
    \caption{Energy Prediction using STPN+convex programming}
    \SetKwInOut{Input}{Input}
    \SetKwInOut{Output}{Output}

    \Input{Training data sets $S, C'_i(i=1,2,3,4)$, depth of $D$}
    \Output{Optimal results $C_i(i=1,2,3,4)$}
    \text{Run all of steps in Algorithm~\ref{STPN}}\;
    \text{Get results by STPN and solve the optimization problem in Eq.~\ref{convex_programming}}\;
    \text{Obtain the optimal results $C_i(i=1,2,3,4)$}\;
\end{algorithm}
\vspace{1cm}
\textbf{Factorial Hidden Markov Model}: Factorial Hidden Markov Model (FHMM) \cite{ZGMJ97} is an extension of Hidden Markov Models that parallelizes multiple Markov models in a distributed manner, and performs some task--related inference to arrive at predicted observation. The application of such models is done by representing each end--use as a hidden state that is modeled by multinomial distribution using $\mathbb{K}$ discrete values, and then sum each appliance meter's individual independent contribution to the expected observation (i.e., the total expected main meter value). AFAMAP \cite{JZKTJ12} variant of FHMM which includes the trends in the hidden states of FHMM have also been reported to be effective in the disaggregation task. In our application of FHMM, the number of hidden states are the number of testing appliances, while $\mathbb{K}$ = 3 in order to keep the computational requirements low.
%\subsubsection{Combinatorial Optimization} \label{sec:co}

\textbf{Combinatorial Optimization}: Combinatorial optimization (CO)~\cite{BKJV11} algorithm is a heuristic scheme that attempts to minimize the $\ell_1$--norm of the total power at the mains and the sum of the power of the end--uses, given either single or multi--state formulation of the sum. The drawbacks of CO for disaggregation tasks are its sensitivity to transients and degradation with increasing number of devices or similarity in device characteristics.\\

We applied the algorithms as available in the non--intrusive load monitoring toolkit~\cite{NJOHWAAM14} with an exact inference~\cite{ZGMJ97} for the FHMM.

\begin{figure}
\centering
\includegraphics[width=0.85\textwidth]{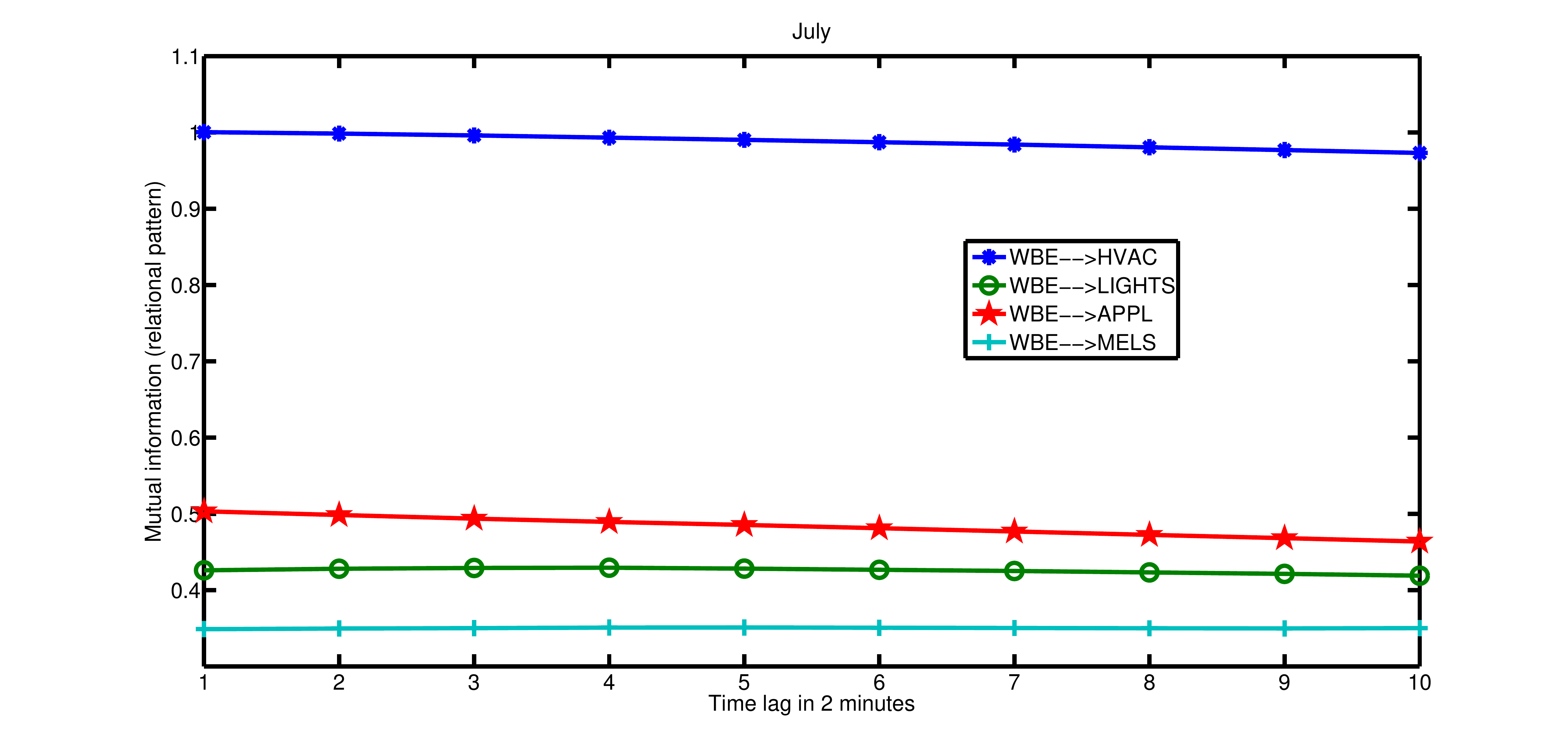}
\caption{\textit{Mutual information between WBE and HVAC, WBE and LIGHTS, WBE and APPL, and WBE and MELS with the increment of time lag of 2 minutes in July, 2010}}\label{MI_1}
\end{figure}

\begin{figure}
\centering
\includegraphics[width=0.85\textwidth]{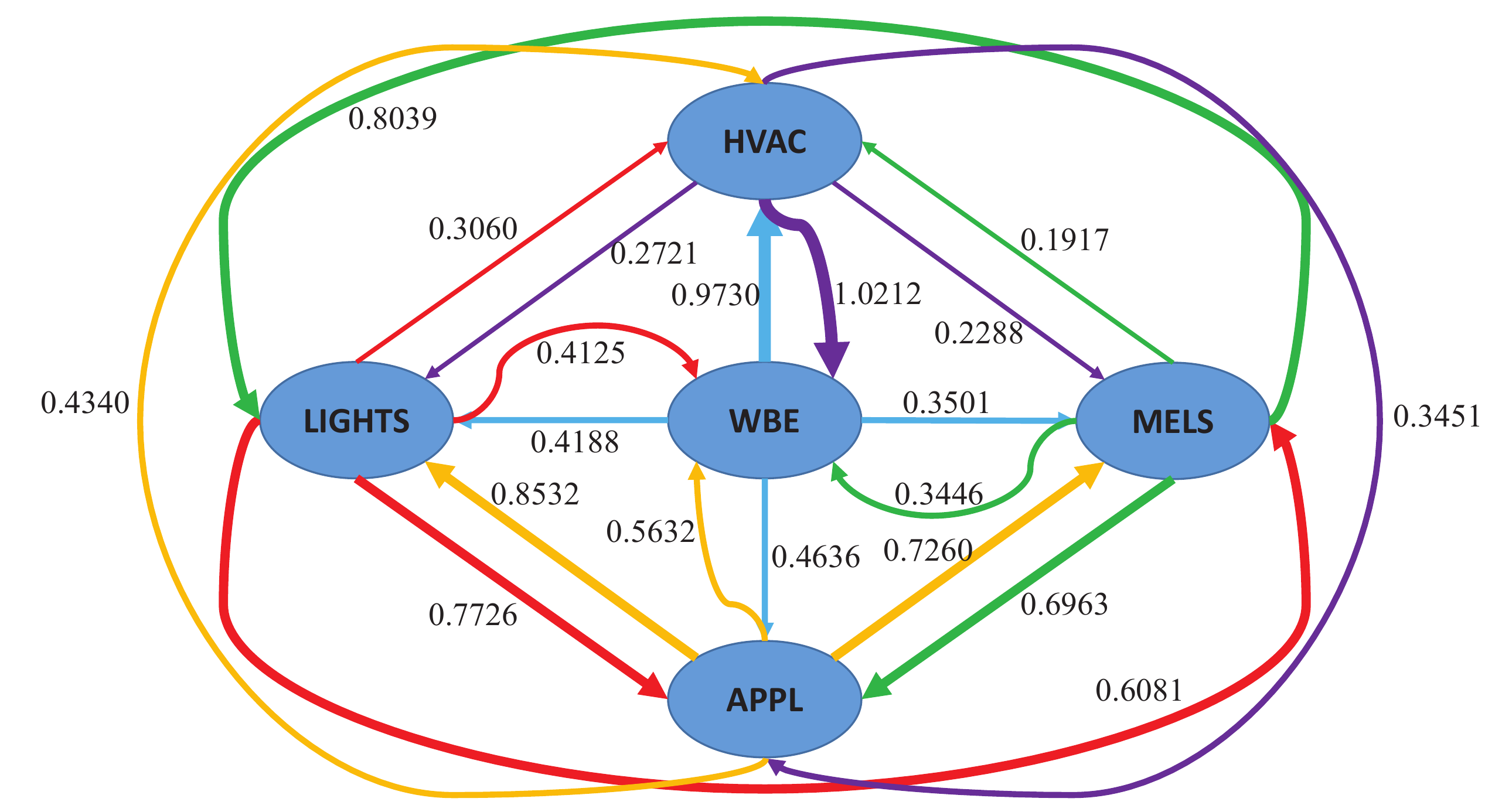}
\caption{\textit{STPN using variables, WBE, HVAC, LIGHTS, APPL and MELS in July}}\label{RP_network}
\end{figure}

\subsection{Results and Discussion}
For validation of the proposed energy prediction approach, two months, i.e., April and July, are selected to study the prediction performance accordingly. As the Building America 2010 data set has 1 hour sampling frequency and three weeks data is for training, such scale of data may not meet the requirement of data size for the construction of STPN. Building up STPN with not enough amount of data may result in the poor accuracy of causal dependencies between different variables. Therefore, a data reprocessing technique, i.e., upsampling is applied in this case and the upsampling fold is 30 such that the sampling frequency for the data set is 2 minutes.

First, we study the causal dependencies among these five variables by computing the mutual information. Figure~\ref{MI_1} shows the variation of mutual information with respect to time lag in 2 minutes for addressing the temporal characteristics. The depth of x$D$Markov machine is still 1 such that the current symbol of any part of HVAC, LIGHTS, APPL and MELS depends only on the past one symbol of WBE. Different from the wind turbine systems, the causal dependencies between WBE and the other four load components have decreased little with an increase of time lag, which reflects that using WBE to predict other parts of energy consumption is temporally robust. However, it also shows that the causal dependency between WBE and HVAC in July is the maximum compared with those between WBE and other load components (i.e., LIGHTS, APPL, and MELS) such that the prediction of HVAC using WBE yields the best accuracy.

The results in Figure~\ref{RP_network} show the causal dependencies quantified by mutual information among all of five variables. It can be observed that the causal dependency between HVAC and APPL is larger than that between HVAC and MELS as well as that between HVAC and LIGHTS. While the relations among LIGHTS, APPL and MELS can be seen to be quite significant due to the causal dependencies obtained in this context. In summary, this relational pattern network captures temporal interactions between different end uses that can be an effective technical tool for energy disaggregation.

\begin{figure}
\centering
\subfigure[]{\includegraphics[width=0.95\textwidth]{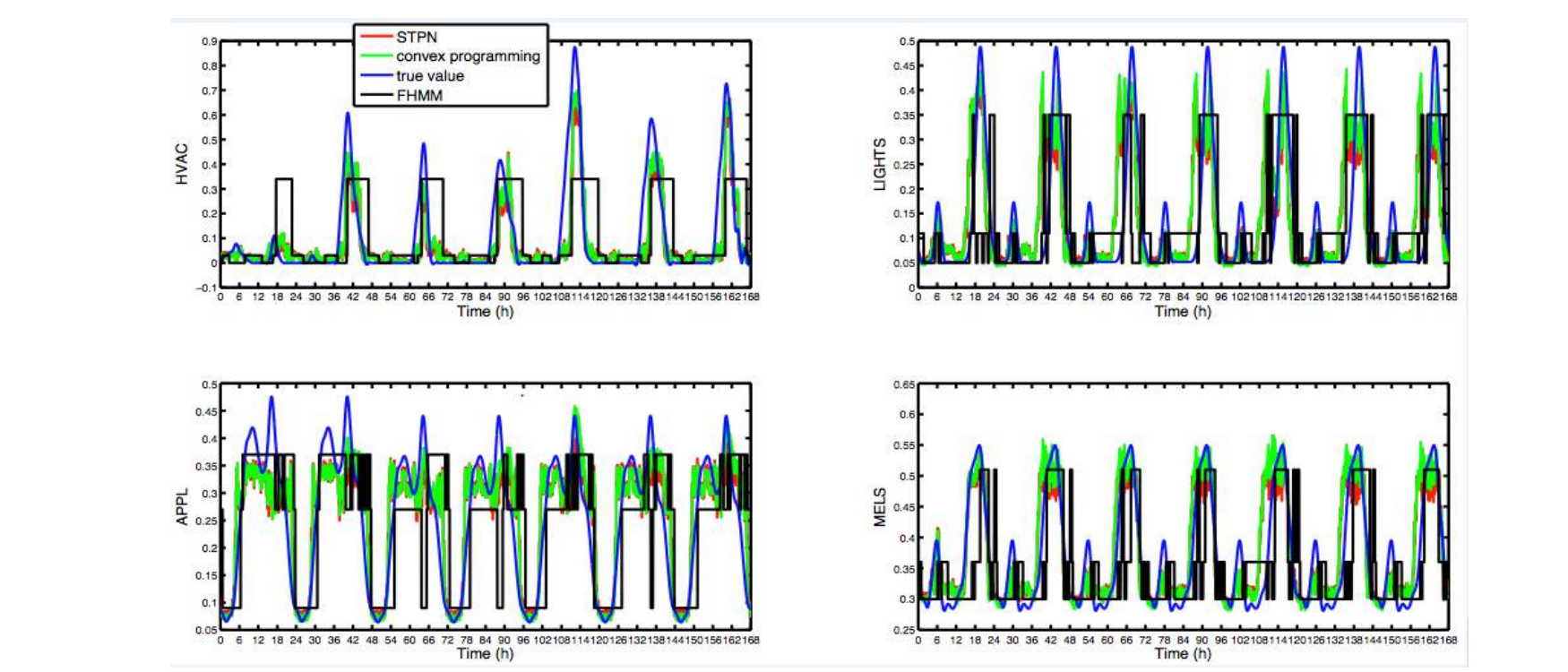}}
\subfigure[]{\includegraphics[width=0.95\textwidth]{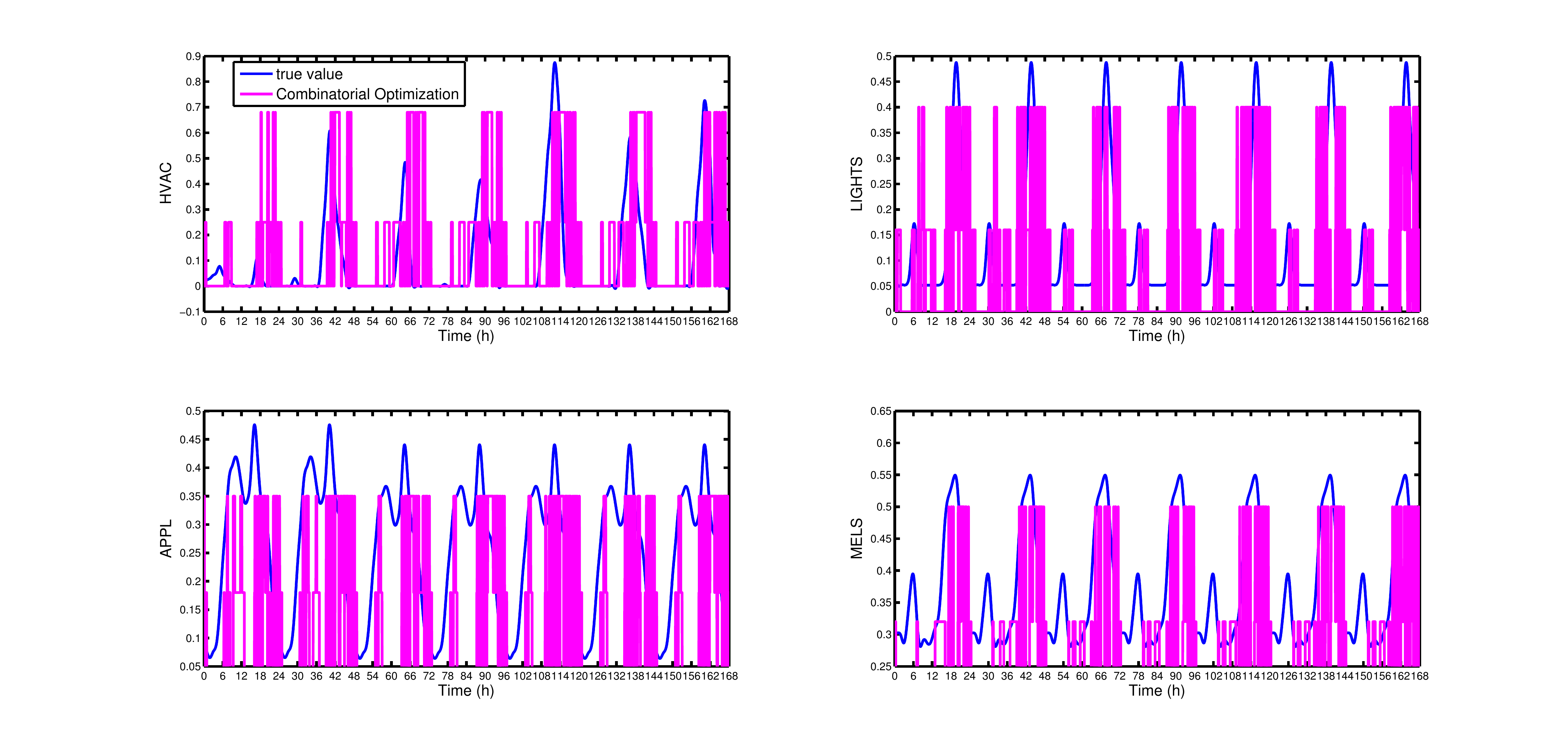}}
\caption{\textit{Energy prediction of HVAC, LIGHTS, APPL, and MELS in April 2010 using STPN, STPN+convex programming, FHMM, and CO separately shown in (b) for better visualization}}\label{Figure14:1}
\end{figure}

\begin{figure}
\centering
\includegraphics[width=0.75\textwidth]{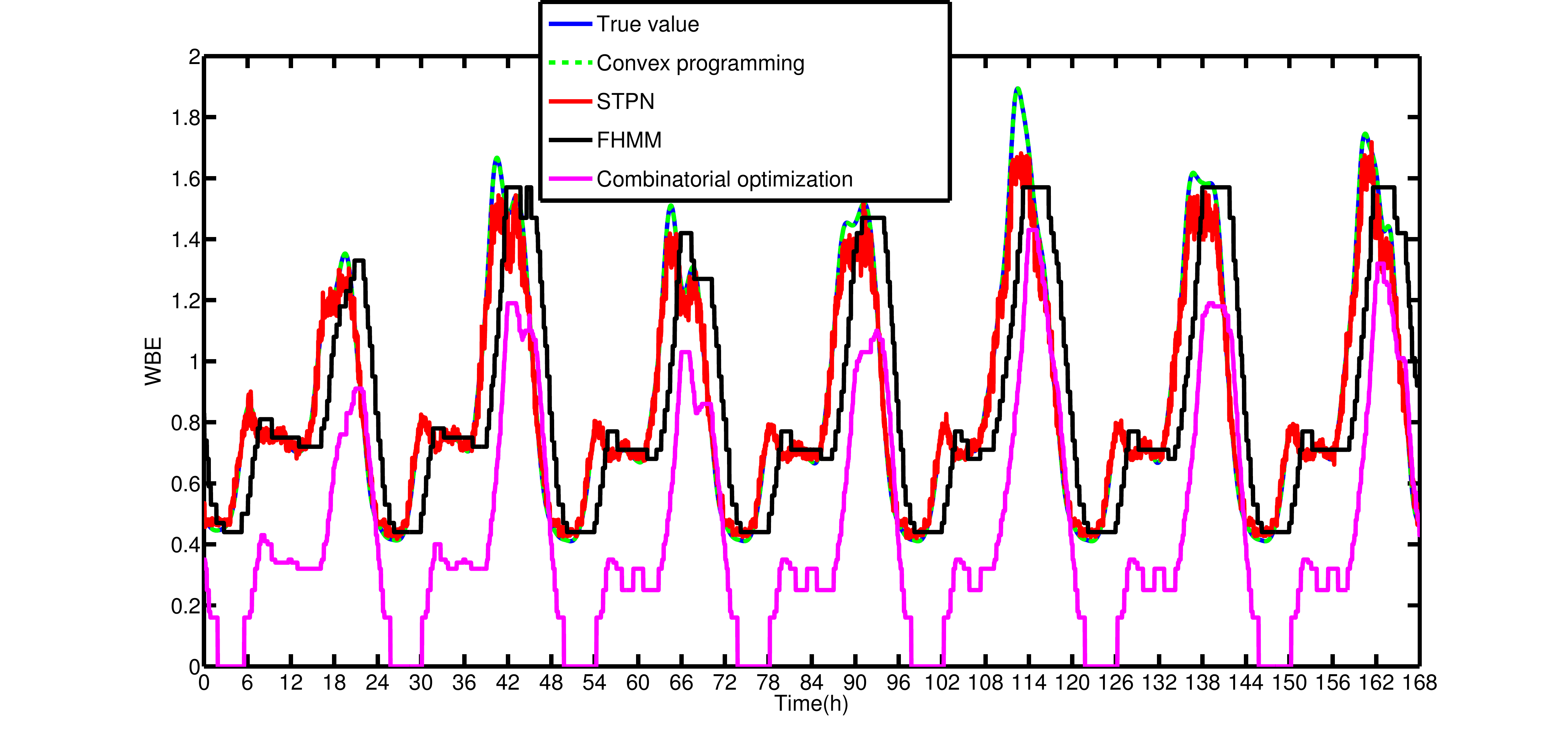}
\caption{\textit{Calculated WBE from disaggregated energy values in April 2010 using STPN, STPN+convex programming, FHMM and CO}}\label{Figure15:1}
\end{figure}

\begin{figure}
\centering
\subfigure[]{\includegraphics[width=0.95\textwidth]{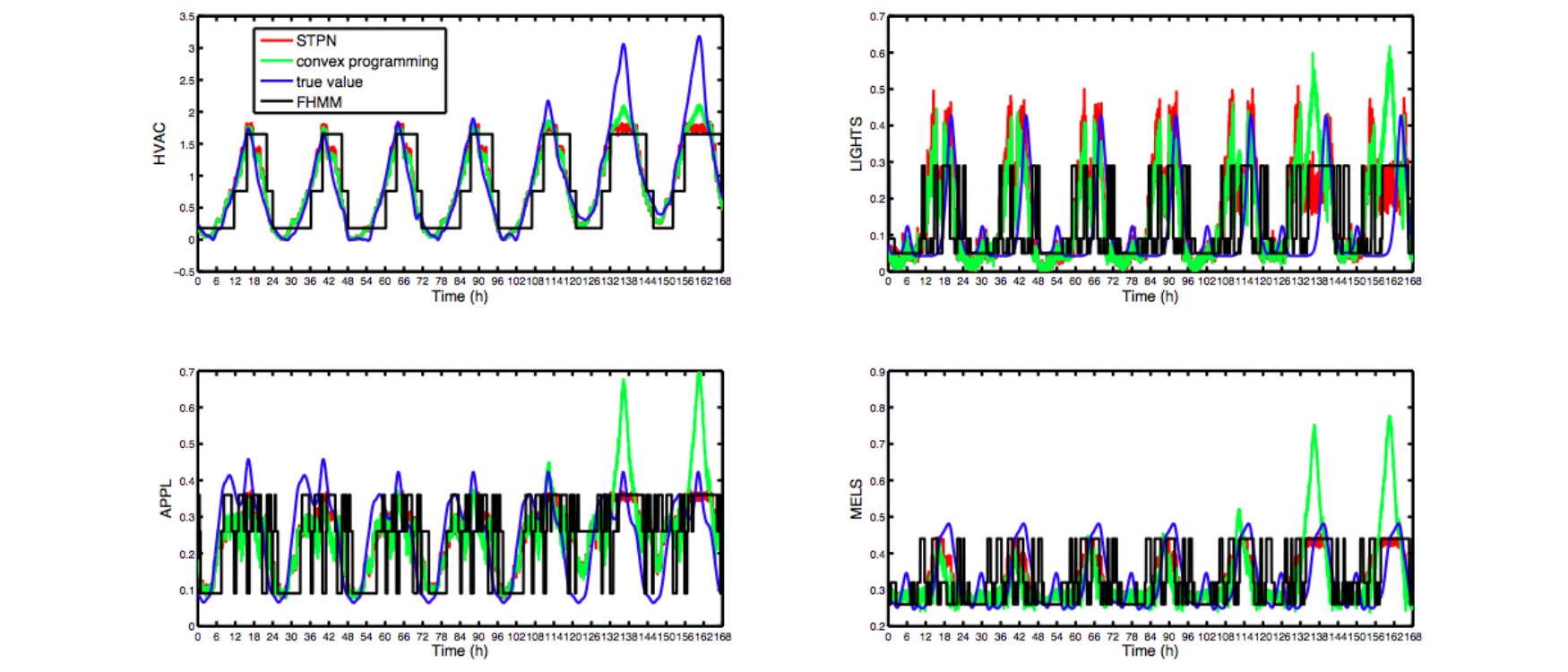}}
\subfigure[]{\includegraphics[width=0.95\textwidth]{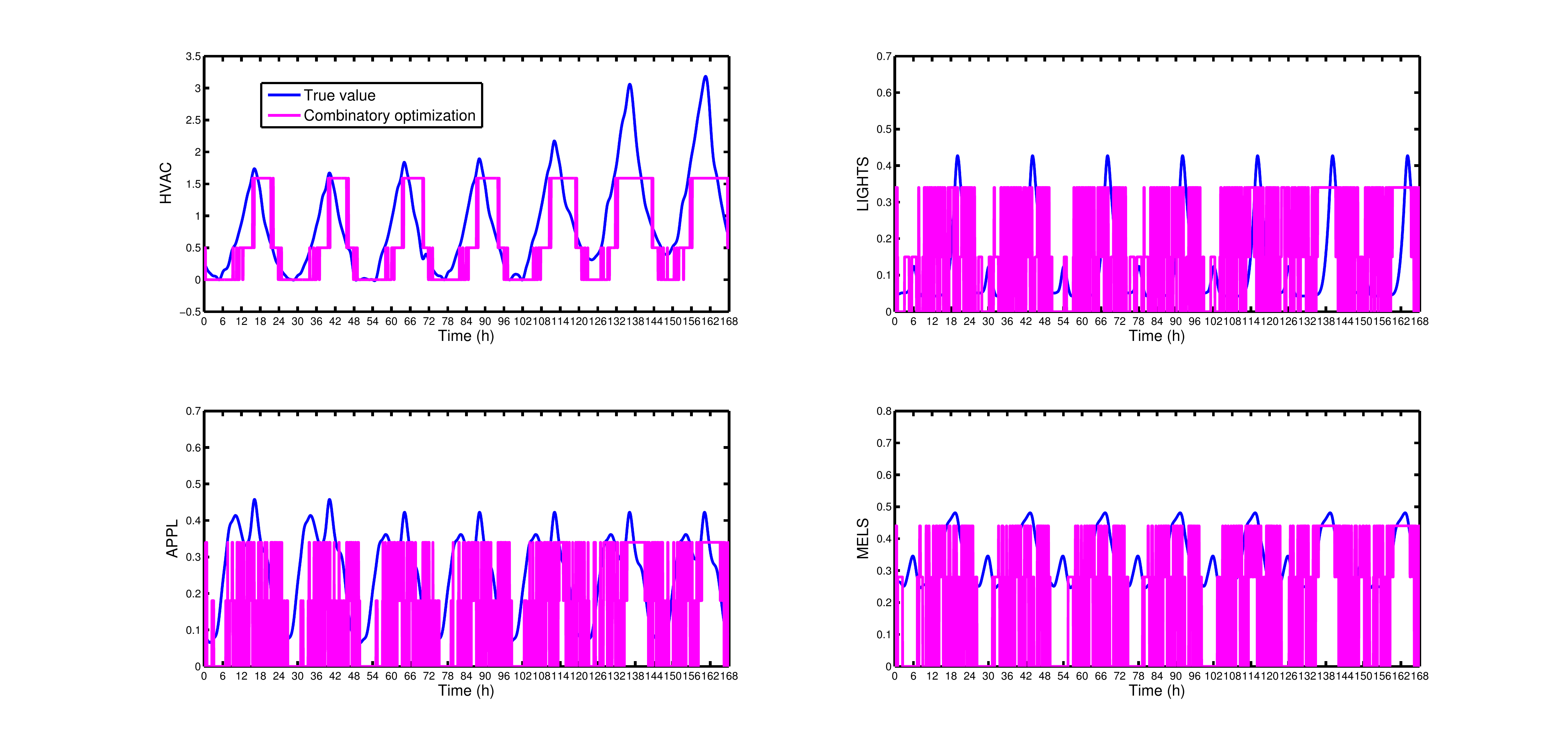}}
\caption{\textit{Energy prediction of HVAC, LIGHTS, APPL, and MELS in July 2010 using STPN, STPN+convex programming, FHMM, and CO separately shown in (b) for better visualization}}\label{Figure16:1}
\end{figure}

Figure~\ref{Figure14:1} shows the energy disaggregation of HVAC, LIGHTS, APPL and MELS using STPN and STPN+convex programming in April. In this month, the energy consumption of HVAC is most significant such that it accounts for the largest percentage of WBE. A strong prediction capabilities of STPN can be observed from the plots and based on that the STPN+convex programming is able to improve STPN performance, which is attributed to the constraint imposed in the convex programming. It can also be seen from Figure~\ref{Figure15:1} that the total energy consumption by STPN without convex programming is worse than STPN+convex programming results and the optimal disaggregation appears to be achieved. However, the prediction performance for APPL and LIGHTS is slightly worse than HVAC and MELS because they account for a lower percentage of WBE, which is also evident as suggested by Figure~\ref{Figure16:1}.

\begin{figure}
\centering
\includegraphics[width=0.75\textwidth]{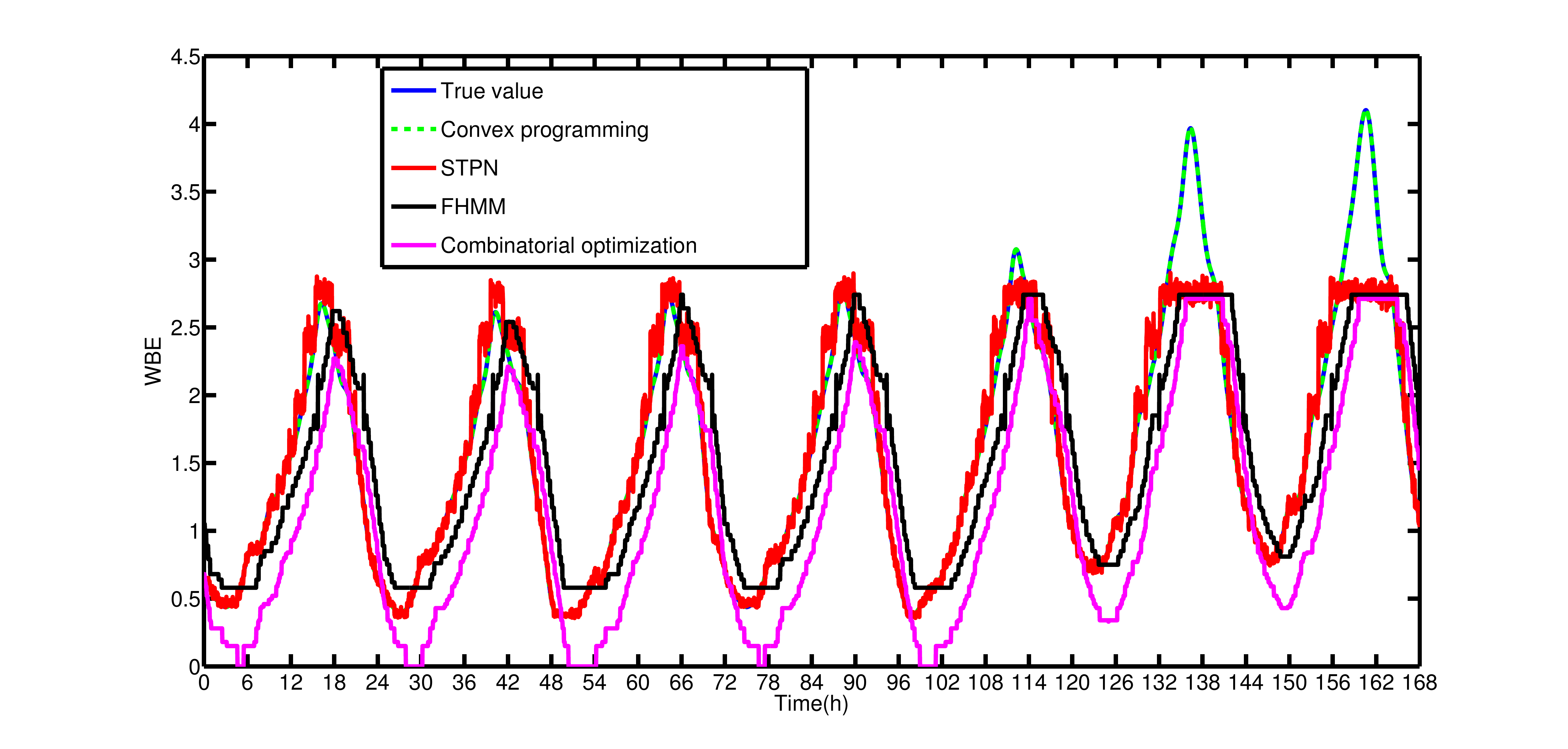}
\caption{\textit{Calculated WBE from disaggregated energy values in July 2010 using STPN, STPN+convex programming, FHMM and CO}}\label{Figure17:1}
\end{figure}

\begin{figure}
\centering
\includegraphics[width=0.95\textwidth]{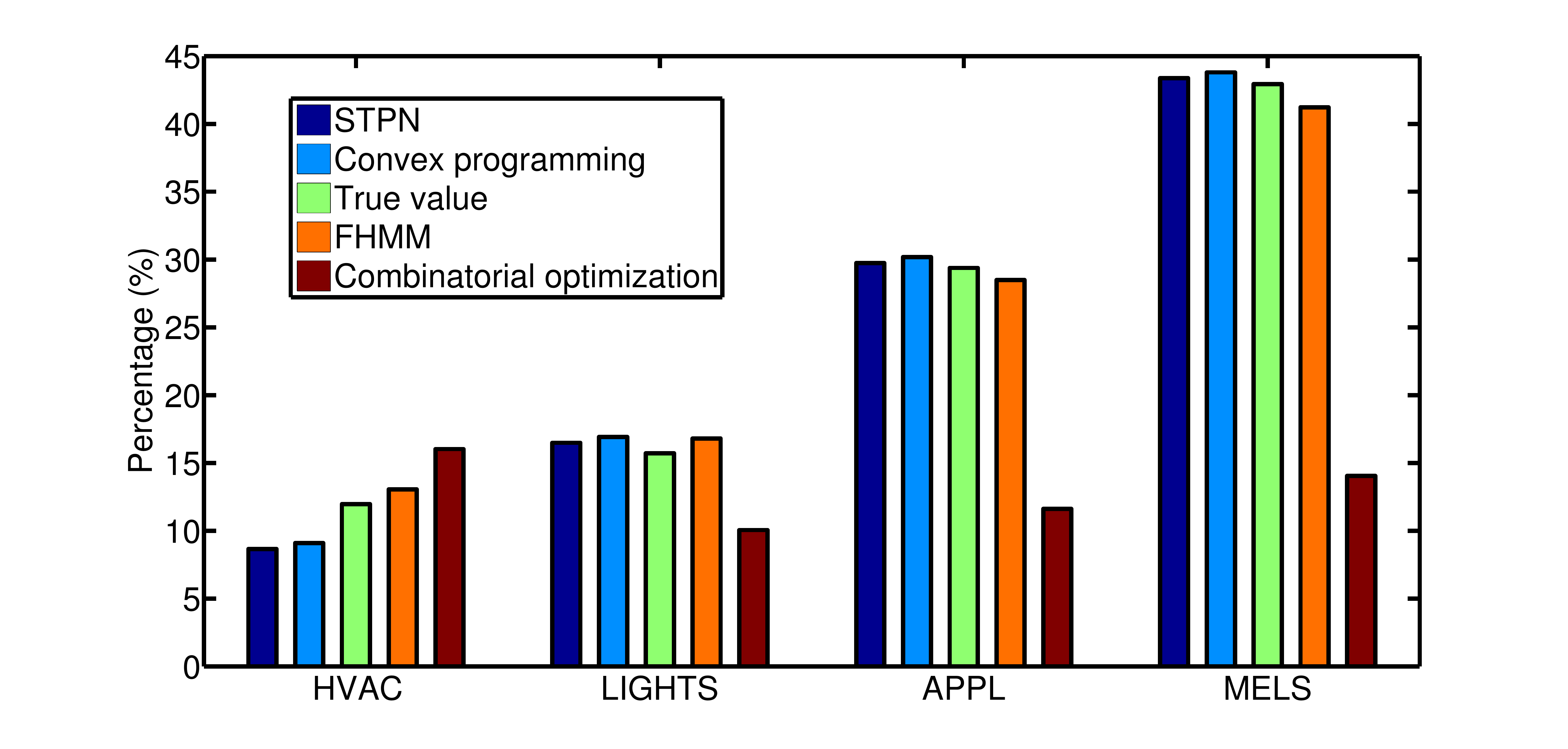}
\caption{\textit{Energy prediction difference of HVAC, LIGHTS, APPL, and MELS in April 2010 among STPN, STPN+convex programming, FHMM and CO}}\label{Figure18:1}
\end{figure}

\begin{figure}
\centering
\includegraphics[width=0.95\textwidth]{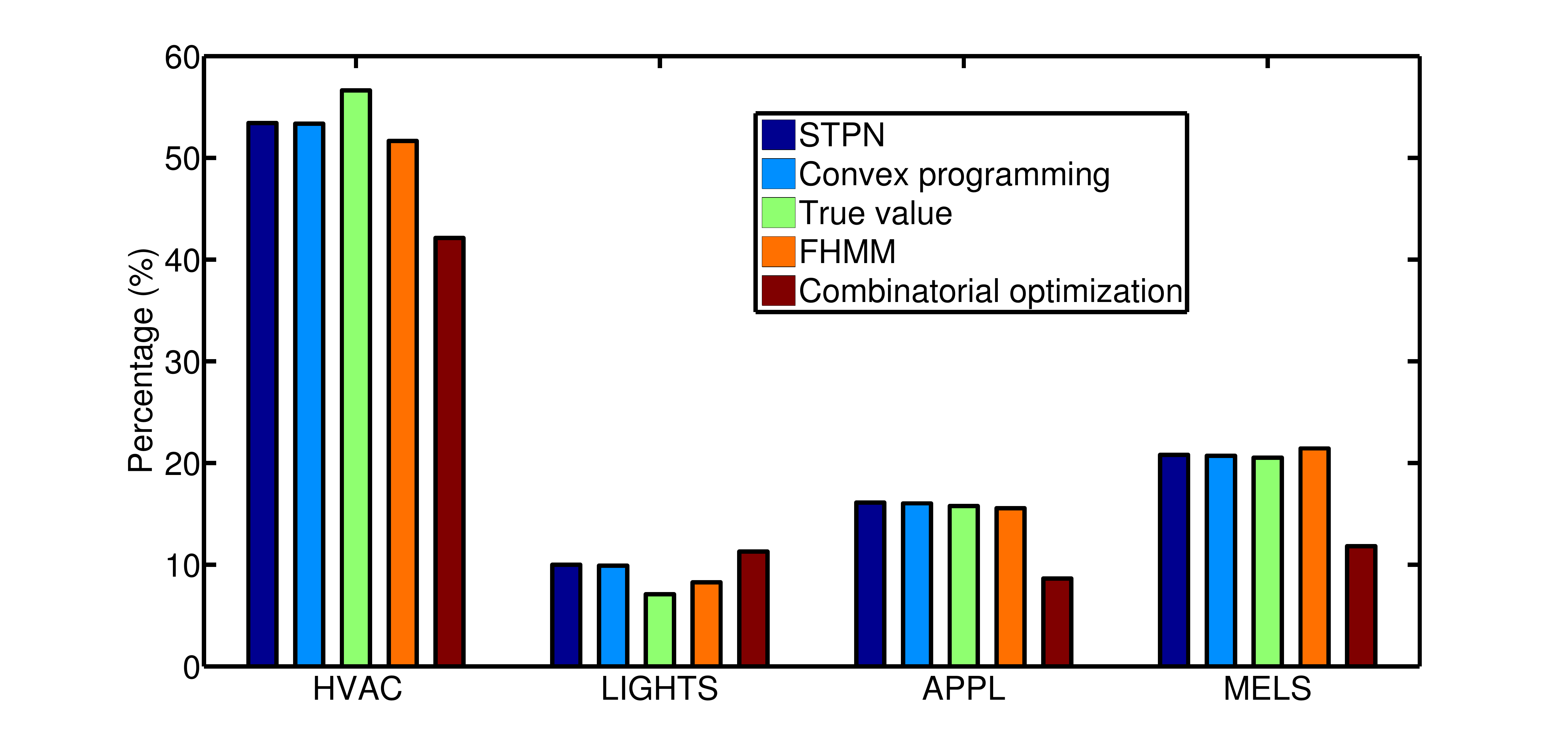}
\caption{\textit{Energy prediction difference of HVAC, LIGHTS, APPL, and MELS in July 2010 among STPN, STPN+convex programming, FHMM and CO}}\label{Figure19:1}
\end{figure}

Therefore, it can be implied that for energy disaggregation the more accurate prediction can be achieved when one load component (i.e., HVAC, LIGHTS, APPL, and MELS) accounts for a more significant percentage of WBE. It is seen from Figure~\ref{Figure16:1} that the prediction for the last two days in the fourth week is worse though it is able to catch the  trend, which may be attributed to the fact that on those two days some transient external factors, such as weather and occupancy, affect the energy consumption. A similar observation can be made from Figure~\ref{Figure17:1} that the optimal disaggregation can be made via STPN+convex programming. For a direct visual inspection of the prediction capability difference, Figure~\ref{Figure18:1} and Figure~\ref{Figure19:1} reveal that STPN+convex programming outperforms STPN alone as for each part the energy consumption is predicted optimally. The fact that these two plots show an energy prediction difference by STPN or STPN+convex programming of less than 5\% demonstrates efficacy and effectiveness of the proposed framework.

%\begin{figure}
%\centering
%\includegraphics[width=0.95\textwidth]{Figures/separate_11_messy_april.eps}
%\caption{\textit{Energy prediction of HVAC, LIGHTS, APPL, and MELS in April 2010 using STPN, convex programming, FHMM, and Combinatorial Optimization}}\label{Figure14:11}
%\end{figure}

%\begin{figure}
%\centering
%\includegraphics[width=0.95\textwidth]{Figures/separate_1_messy_july.eps}
%\caption{\textit{Energy prediction of HVAC, LIGHTS, APPL, and MELS in July 2010 using STPN, convex programming, FHMM, and Combinatorial Optimization}}\label{Figure16:11}
%\end{figure}

To see the comparison between the proposed method and the current state-of-the-art techniques in literature, in this context we compare the STPN and STPN+convex programming method to FHMM and CO. However, for obtaining enough accuracy of prediction results, the data set is as well upsampled for FHMM with upsampling fold being 1200. Thus the sampling frequency becomes 3 sec accordingly and the number of states used is 3. The energy disaggregation results from Figure~\ref{Figure14:1} show that both FHMM and CO perform worse than the proposed method although the predicted WBE in Figure~\ref{Figure15:1} looks quite promising. It is because FHMM cannot predict the transient peaks appearing as quite well as the proposed method and CO is unable to disaggregate the load component well. The very similar conclusion is made as well for the month of July. From Figure~\ref{Figure16:1} it is observed that when the energy curves are more oscillatory, the proposed method is able to outperform FHMM and CO. It can be suggested both from Figures~\ref{Figure15:1} and~\ref{Figure17:1} that the proposed STPN and STPN+convex programming present better energy prediction in terms of WBE. Results in Figure~\ref{Figure18:1}, ~\ref{Figure19:1} and Table~\ref{table2} quantitatively present the difference among the proposed method (STPN, STPN+convex programming), FHMM, and combinatorial optimization method. It strengthens the conclusion that using STPN and STPN+convex programming yield quite encouraging and promising disaggregation results in NILM. Hence, the comparison among the proposed method and FHMM, combinatorial optimization indicates the effectiveness of the STPN-based energy prediction scheme as an important tool to deal with energy prediction. We also remark on the computational efficiency on the proposed method, FHMM, and CO.

\begin{table}[h!]
  \centering
  \caption{Computational information for different methods in April}
  \label{table2}
  \begin{tabular}{cccc}
    \toprule
    Method & Time (s) & Memory (MB) & Accuracy (MSE)\\
    \midrule
    STPN & 28.74 & 962 & 0.0072\\
    STPN+convex programming & 369.64 & 2756 & 0.0070\\
    FHMM & 38.10 & 798.67 & 0.0163\\
    CO & 11.25 & 769.37 & 0.0564\\
    \bottomrule
  \end{tabular}
\end{table}

\begin{remark}
In this case we also consider the computational time, memory along with accuracy (MSE) in order to compare the performance of different methods. FHMM and combinatorial optimization methods were implemented in ipython notebook for the NILM toolkit (NILMTK) while STPN and STPN+convex programming in the MATLAB environment and CVX package~\cite{grant2008cvx}. The results in Table~\ref{table2} show that STPN can spend less time than FHMM while more memory is required as the number of states for STPN is more than FHMM in this case. STPN+convex programming approach needs more computational time and memory to run the whole process due to the optimizing iterations. FHMM and CO use less memory compared to the proposed schemes. However, in terms of accuracy, the STPN outperforms FHMM and CO approaches as shown in Table~\ref{table2}. The MSE of FHMM is more than two times as that of STPN. Moreover, STPN+convex programming is able to improve the accuracy obtained from the STPN framework. In summary, the energy prediction method based on the STPN framework may be an effective way in the applications of energy prediction. Note, the FHMM and the CO codes used here are part of a well--optimized toolbox and we expect that similar code and platform optimization can bring our proposed methods to a comparable level in terms of memory and time complexity.
\end{remark}
%\begin{figure}
%\centering
% \includegraphics[width=0.5\textwidth]{Figures/gp1.eps}
%\caption{Zone temperature prediction (mean and uncertainty bound) using Gaussian Process}
%\label{fig:gp}
%\end{figure}

\section{Conclusions and Future Work}\label{sec:con}
This paper presents a novel data-driven framework, spatiotemporal pattern network (STPN) to predict energy consumption for both supply side and demand side energy systems. While symbolic dynamic filtering performs the discretization and symbolization of continuous domain data for data level fusion of different variables in a dynamic system, a $D$-Markov machine is able to capture its temporal characteristics. This work establishes another PFSA, called x$D$-Markov machine, for addressing the issue of how to capture the causal dependencies between two time-series in this work. Moreover, for the quantification of causal dependencies, a mutual information based metric is applied in this regard. Prediction based on the STPN framework is proposed using expectation from symbolic domain to symbolic and continuous domain.

The proposed scheme is validated by two case studies, wind turbine power prediction (supply side energy systems) and non-intrusive load monitoring (demand side energy systems). For wind power prediction, the primary observation made in this paper is that the proposed STPN models can capture the salient spatiotemporal features and it is demonstrated that causal dependencies decrease with an increase in both spatial distances and temporal lags as intuitively expected. Based on such observation, the power prediction for a wind turbine is performed by using the observation from another wind turbine with a high degree of accuracy. For non-intrusive load monitoring, energy disaggregation performance of the proposed STPN framework with and without a convex programming step is evaluated. While the STPN scheme shows that each part of disaggregated energy can be predicted significantly better than state-of-the-art techniques such as FHMM and combinatorial optimization, a convex programming approach based on STPN is able to improve the prediction performance to achieve a further optimized disaggregation involving the constraint -- disaggregated energy values should sum up to the total energy usage.

While current efforts are focusing on applying the proposed techniques on real data and problems, some of the other future research directions include:
\begin{enumerate}
\item For wind power prediction -- Impact analysis of other physical variables, e.g., wind direction on model quality for wind power prediction;
\item For energy disaggregation -- Joint state prediction by taking multiple variables into account for energy disaggregation;
\item For energy disaggregation -- Weighted factor and penalty term analysis in convex optimization for energy disaggregation.
\end{enumerate}

\section*{Acknowledgement}
This work was supported by the National Science Foundation under Grant No. CNS-1464279.
%This paper has proposed the use of a novel spatiotemporal pattern network (STPN) framework to capture the interaction characteristics between multiple wind turbines. While the discretization and symbolization steps of SDF performs data level fusion of wind power and wind speed for a single wind turbine system, a $D$-Markov machine captures its stationary temporal dynamics. Causal dependency between two turbines is modeled using a variant called the x$D$-Markov machine. Moreover, the causal dependency is quantified by a mutual information based metric. The proposed scheme is validated using the Western Wind Integration data set from NREL. The primary observation made is that the Markov machines and the mutual information based causality metric are able to capture both temporal and spatial characteristics as causality decreases with increase in both temporal lags and spatial distances. The STPN scheme is further evaluated using prediction of wind power production by one wind turbine using observed symbol sequence from another turbine. Some of the future research directions currently being pursued are:
%\begin{enumerate}
%\item Impact analysis of other physical variables, e.g., wind direction on model quality;
%\item Systematic (short and long term) farm-wide wind power prediction using STPN;
%\item Simulating ``what if" scenarios using STPN for distributed optimization of farm-wide wind power production.
%\end{enumerate}

\bibliography{AppliedEnergy_WindEnergy}

\end{document}